\documentclass[twoside,11pt]{article}

\usepackage{jair}
\usepackage{theapa}
\usepackage{amsmath}
\usepackage{amssymb}
\usepackage{amsfonts}
\usepackage[dvips]{graphicx}
\usepackage[dvips]{color}
\usepackage{epsf}
\usepackage{epsfig}
\usepackage{url}
\usepackage{psfrag}
\usepackage{multirow}

\newcommand{\me}[1]{\mbox{\it #1}}
\newcommand{\keyw}[1]{{\bf #1}}

\DeclareMathOperator{\mgu}{mgu}
    
\DeclareMathOperator{\hb}{hb}

\DeclareMathOperator{\dom}{D}

\DeclareMathOperator{\vars}{vars}

\DeclareMathOperator{\cl}{\mathtt{T}}  
\DeclareMathOperator{\gcl}{gcl}

\newtheorem{exam}{Example}
\newtheorem{defn}{Definition}
\newtheorem{thm}{Theorem}

\jairheading{25}{2006}{425--456}{12/04}{4/06}

\ShortHeadings{Logical Hidden Markov Models}
{Kersting, De Raedt, \& Raiko}

\firstpageno{425}

\begin{document}

\title{Logical Hidden Markov Models}

\author{\name Kristian Kersting \email kersting@informatik.uni-freiburg.de\\
	\name Luc De Raedt \email deraedt@informatik.uni-freiburg.de\\
	\addr Institute for Computer Science\\
              Albert-Ludwigs-Universit{\"a}t Freiburg\\ 
              Georges-Koehler-Allee 079\\ 
              D-79110 Freiburg, Germany
	\AND 
	\name Tapani Raiko \email tapani.raiko@hut.fi\\
        \addr Laboratory of Computer and Information Science\\
              Helsinki University of Technology\\
              P.O. Box 5400\\
              FIN-02015 HUT, Finland}
\maketitle

\begin{abstract}
Logical hidden Markov models (LOHMMs) upgrade 
traditional hidden Markov models to deal with
sequences of structured symbols in the form of logical atoms, rather than flat characters.

This note formally introduces LOHMMs and presents solutions to the three
central inference problems for LOHMMs: evaluation, most likely hidden state sequence
and parameter estimation. The resulting representation and algorithms 
are experimentally evaluated on problems from the domain of bioinformatics.
\end{abstract}

\section{Introduction}
Hidden Markov models (HMMs)~\cite{rabiner86introduction} are extremely
popular for analyzing sequential data. Application areas include
computational biology, user modelling, speech recognition, empirical
natural language processing, and robotics.  Despite their successes,
HMMs have a major weakness: they handle only sequences of flat, i.e.,
unstructured symbols. Yet, in many applications the symbols
occurring in sequences are structured.  Consider, e.g., sequences of
\textsc{UNIX} commands, which may have parameters such as ${\mathtt{emacs \ lohmms.tex},\mathtt{ls},}$
${\mathtt{latex \ lohmms.tex},\ldots}$%
Thus, commands are essentially
structured. 
Tasks that have been considered for
\textsc{UNIX} command sequences include the prediction of the next
command in the sequence~\cite{DavisonHirsh:98}, the classification
of a command sequence in a user
category~\cite{KorvemakerGreiner:00,JacobsBlockeel:01}, and  
anomaly detection~\cite{Lane:99}. 
Traditional HMMs cannot easily deal with this type of structured
sequences. 
Indeed, applying HMMs requires either 1)
ignoring the structure of the commands (i.e., the parameters), or 2)
taking all possible parameters explicitly into account. The former
approach results in a serious information loss; the latter leads to a
combinatorial explosion in the number of symbols and parameters of the HMM and 
as a consequence inhibits generalization.

The above sketched problem with HMMs is akin to the problem of dealing with
structured examples in traditional machine learning algorithms as 
studied in the fields of inductive logic programming~\cite{MuggletonDeRaedt:94}
and multi-relational learning~\cite{DzeroskiLavrac:01}. 
In this paper, we propose an (inductive) logic programming framework, Logical HMMs (LOHMMs),
that upgrades HMMs to deal with structure.
The key idea underlying LOHMMs is to employ 
logical atoms as structured (output and state) symbols.
Using logical atoms, the above \textsc{UNIX} command sequence can be represented as 
$\mathtt{emacs(lohmms.tex)},\mathtt{ls},\mathtt{latex(lohmms.tex)},\ldots$ 
There are two important motivations for using logical atoms at the symbol level.
First, {\em variables} in
the atoms allow one to make abstraction of specific symbols.
E.g.,\ the logical atom $\mathtt{emacs(X,tex)}$  represents all files $\mathtt{X}$ that a \LaTeX \ user 
$\mathtt{tex}$
could edit using $\mathtt{emacs}$. 
Second, {\em unification} allows one to share information among %hidden 
states. 
E.g.,\ the sequence $\mathtt{emacs(X,tex)},\mathtt{latex(X,tex)}$ 
denotes that the same file is used as an argument for both Emacs and \LaTeX.

The paper is organized as follows. 
After reviewing the logical preliminaries, % in Section~\ref{sec:preliminaries},
we introduce LOHMMs and define their semantics in Section~\ref{sec:lohmms};
in Section~\ref{sec:inf}, we upgrade the basic HMM inference algorithms 
for use in LOHMMs; we investigate the benefits of LOHMMs in Section~\ref{sec:bene}:
we show that LOHMMs are strictly more
expressive than HMMs, that they can be --- by design --- %at least 
an order of magnitude smaller
than their corresponding propositional instantiations, and that unification can yield models, which better fit the data.
In Section~\ref{sec:eval}, we empirically investigate the benefits of LOHMMs 
on real world data. Before concluding, we discuss related work in Section~\ref{sec:related}.
Proofs of all theorems can be found in the Appendix.

\section{Logical Preliminaries}\label{sec:preliminaries}
A {\em first-order alphabet} $\Sigma$ is a set of relation symbols $\mathtt{r}$
with arity $m\geq 0$, written $\mathtt{r}/m$, and a set of functor symbols $\mathtt{f}$
with arity $n\geq 0$, written $\mathtt{f}/n$.  If $n=0$ then $\mathtt{f}$ is called a
constant, if $m=0$ then $\mathtt{p}$ is called a propositional variable. (We
assume that at least one constant is given.)  An {\it atom}
$\mathtt{r(t}_1\mathtt{,\ldots ,t}_n\mathtt{)}$ is a relation symbol $\mathtt{r}$ followed by a bracketed
$n$-tuple of terms $\mathtt{t}_i$.  A {\it term} $\mathtt{t}$ is a variable $\mathtt{V}$ or a
functor symbol $\mathtt{f(t}_1\mathtt{,\ldots ,t}_k\mathtt{)}$ immediately followed by a bracketed
$k$-tuple of terms $\mathtt{t}_i$. Variables will be written in upper-case, and constant, functor and predicate
symbols lower-case. The symbol $\_$ will denote anonymous variables which are 
read and treated as distinct, new variables each time they are encountered. 
An {\it iterative clause} is a formula of the form
$\mathtt{H\leftarrow B}$ where $\mathtt{H}$ (called {\it head}) and $\mathtt{B}$ (called {\it body}) are logical atoms. 
A substitution $\theta=\{\mathtt{V}_1/\mathtt{t}_1,\ldots ,\mathtt{V}_n/\mathtt{t}_n\}$, e.g.
$\{\mathtt{X}/\mathtt{tex}\}$, is an assignment of terms $\mathtt{t}_i$ to variables $\mathtt{V}_i$.
Applying a substitution $\sigma$ to a term, atom or clause $\mathtt{e}$ yields
the instantiated term, atom, or clause $\mathtt{e}\sigma$ where all occurrences
of the variables $\mathtt{V}_i$ are simultaneously replaced by the term $\mathtt{t}_i$,
e.g. $\mathtt{ls(X)\leftarrow emacs(F,X)}\{\mathtt{X}/\mathtt{tex}\}$ yields
$\mathtt{ls(tex)\leftarrow emacs(F, tex)}$. 
A substitution
$\sigma$ is called a {\it unifier} for a finite set $S$ of atoms if
$S\sigma$ is singleton. A unifier $\theta$ for $S$ is called a 
{\it most general unifier} (MGU) for $S$ if, for each unifier $\sigma$ of
$S$, there exists a substitution $\gamma$ such that 
$\sigma=\theta\gamma$. 
A term, atom or clause
$\mathtt{E}$ is called {\it ground} when it contains no variables, 
i.e., $\me{vars}(\mathtt{E})=\emptyset$.  The {\it Herbrand base} of $\Sigma$,
denoted as $\hb_{\Sigma}$, is the set of all ground atoms constructed
with the predicate and functor symbols in $\Sigma$.  The set
$G_{\Sigma}(\mathtt{A})$ of an atom $\mathtt{A}$
consists of all ground atoms $\mathtt{A}\theta $ that belong to $\hb_{\Sigma}$.

\section{Logical Hidden Markov Models}\label{sec:lohmms}
The logical component of a traditional HMM corresponds to a {\em Mealy
machine}~\cite{HopcroftUllman:79}, i.e., a finite
state machine where the output symbols are associated with transitions. 
This is essentially a propositional representation
because the symbols used to represent states and output symbols are flat, i.e. not structured.
The key idea underlying LOHMMs 
is to replace these flat symbols by abstract symbols. An abstract symbol $\mathtt{A}$ is --- by definition --- a logical atom.
It is abstract in that it represents 
the set of all ground, i.e., variable-free atoms of $\mathtt{A}$ over the alphabet $\Sigma$, denoted by $G_\Sigma(\mathtt{A})$.
Ground atoms then play the role of the traditional symbols
used in a HMMs. 
\begin{exam}
Consider the alphabet $\Sigma_1$ which has as constant symbols $\mathtt{tex}$, $\mathtt{dvi}$,
$\mathtt{hmm1}$, and $\mathtt{lohmm1}$, and as relation symbols $\mathtt{emacs}/2$, $\mathtt{ls}/1$, $\mathtt{xdvi}/1$, 
$\mathtt{latex}/2$.
Then the atom $\mathtt{emacs(File,tex)}$ represents the set $\{\mathtt{emacs(hmm1,tex)},\mathtt{emacs(lohmm1,tex)}\}$.
We 
assume that the alphabet is typed to avoid useless instantiations
such as $\mathtt{emacs(tex,tex)}$).
\end{exam}
The use of atoms instead of flat symbols allows us to analyze logical and structured sequences such as 
$\mathtt{emacs(hmm1,tex)}, \mathtt{latex(hmm1,tex)}, \mathtt{xdvi(hmm1,dvi)}$. 
\begin{defn}
\emph{Abstract transition} are expressions of the form
${p:\mathtt{H \ \xleftarrow{\tiny O} \ B}}$ where $p \in [0,1]$,
and $\mathtt{H}$, $\mathtt{B}$
and $\mathtt{O}$ are atoms. All variables are implicitly assumed to be universally quantified, i.e.,
the scope of variables is a single abstract transition.
\end{defn}
The atoms $\mathtt{H}$ and $\mathtt{B}$ represent abstract states
and $\mathtt{O}$ represents an abstract output symbol.
The semantics of an abstract transition ${p:\mathtt{H \ \xleftarrow{\tiny O} \ B}}$ is that if one is in one of the states 
in $G_\Sigma( \mathtt{B})$, say $\mathtt{B}\theta_{\mathtt{B}}$, one will go with probability $p$ to one of the states in $G_\Sigma(\mathtt{H}\theta_{\mathtt{B}})$, say
$\mathtt{H}\theta_{\mathtt{B}}\theta_{\mathtt{H}}$, while emitting a %one of the 
symbol in $G_\Sigma(\mathtt{O}\theta_{\mathtt{B}}\theta_{\mathtt{H}})$, say~$\mathtt{O}\theta_{\mathtt{B}}\theta_{\mathtt{H}}\theta_{\mathtt{O}}$. 
\begin{exam}
Consider $c\equiv{0.8: \mathtt{xdvi(File,dvi) \xleftarrow{\tiny latex(File)}latex(File,tex)}}$.
In general $\mathtt{H}$, $\mathtt{B}$ and $\mathtt{O}$ do not have to share the same predicate. This is only due to the nature 
of our running example. 
Assume now that we are in state $\mathtt{latex(hmm1,tex)}$, i.e. $\theta_{\mathtt{B}} = \{ \mathtt{File}/\mathtt{hmm1} \}$.
Then $c$ 
specifies that there is a probability of 0.8 that the next state will be in $G_{\Sigma_1}(\mathtt{xdvi(hmm1,dvi)}) 
= \{ \mathtt{xdvi(hmm1,dvi)}\}$ ( i.e.,  the probability is $0.8$ that the next state will be 
$\mathtt{xdvi(hmm1,dvi)}$), and that one of the symbols in 
$G_{\Sigma_1}(\mathtt{latex(hmm1)})=\{\mathtt{latex(hmm1)}\}$ ( i.e.,
$\mathtt{latex(hmm1)}$) will be emitted. 
Abstract states might also be more complex such as $\mathtt{latex(file(FileStem,FileExtension),User)}$
\end{exam}
The above example was simple because $\theta_{\mathtt{H}}$ and $\theta_{\mathtt{O}}$
were both empty. The situation becomes more complicated when these substitutions
are not empty. Then,
the resulting state and output symbol sets are not necessarily singletons.
Indeed, for the transition  ${0.8: \mathtt{emacs(File^\prime,dvi) \xleftarrow{\tiny latex(File)} latex(File,tex)}}$ 
the resulting state set would be 
$G_{\Sigma_1}(\mathtt{emacs(File^\prime,dvi)})=\{ \mathtt{emacs(hmm1,tex)}, \mathtt{emacs(lohmm1,tex)}\}$.
Thus the transition is non-deterministic because there
are two possible resulting states. We therefore need a mechanism
to assign probabilities to these possible alternatives.
\begin{defn}
The selection distribution $\mu$ specifies for each abstract state and observation symbol $\mathtt{A}$ over the alphabet $\Sigma$
a distribution $\mu(\cdot\mid \mathtt{A})$ over 
$G_{\Sigma}(\mathtt{A})$.
\end{defn}
To continue our example, let $\mu(\mathtt{emacs(hmm1,tex)} \mid \mathtt{emacs(File^\prime,tex)}) = 0.4$
and $\mu(\mathtt{emacs(lohmm1,tex)} \mid \mathtt{emacs(File^\prime,tex)}) = 0.6$. Then there would be 
a probability of $0.4 \times 0.8 = 0.32 $ that the next state 
is $\mathtt{emacs(hmm1,tex)}$ and of $0.48$ that it is 
$\mathtt{emacs(lohmm1,tex)}$.

Taking $\mu$ into account, 
the meaning of an abstract transition ${p:\mathtt{H \ \xleftarrow{\tiny O} \ B}}$ 
can be summarized as follows.
Let $\mathtt{B}\theta_{\mathtt{B}} \in G_{\Sigma}(\mathtt{B})$, $\mathtt{H}\theta_{\mathtt{B}}\theta_{\mathtt{H}} \in G_{\Sigma}(\mathtt{H}\theta_{\mathtt{B}})$ and $\mathtt{O}\theta_{\mathtt{B}}\theta_{\mathtt{H}}\theta_{\mathtt{O}} \in
G_{\Sigma}(\mathtt{O}\theta_{\mathtt{B}}\theta_{\mathtt{H}})$. Then the model makes a transition from
state $\mathtt{B}\theta_{\mathtt{B}}$ to $\mathtt{H}\theta_{\mathtt{B}}\theta_{\mathtt{H}}$ and emits symbol 
$\mathtt{O}\theta_{\mathtt{B}}\theta_{\mathtt{H}}\theta_{\mathtt{O}}$
with probability
\begin{align}\label{trans_prob}
p\cdot\mu(\mathtt{H}\theta_{\mathtt{B}}\theta_{\mathtt{H}}\mid \mathtt{H}\theta_{\mathtt{B}})\cdot\mu(\mathtt{O}\theta_{\mathtt{B}}\theta_{\mathtt{H}}\theta_{\mathtt{O}}\mid \mathtt{O}\theta_{\mathtt{B}}\theta_{\mathtt{H}}).
\end{align}
To represent $\mu$, any probabilistic representation can - in principle -  be used, e.g.
a Bayesian network or a Markov chain. Throughout the remainder of the present paper, however, we will use  
a {\it na\"\i ve Bayes} approach. More precisely,
we associate to each 
argument of a relation $\mathtt{r}/m$ a finite
domain $\dom_i^{\mathtt{r}/m}$ of constants and a probability distribution $P_i^{\mathtt{r}/m}$ over
$\dom_i^{\mathtt{r}/m}$. Let  $\vars(\mathtt{A})=\{\mathtt{V}_{1},\ldots, \mathtt{V}_{l}\}$ be the variables occurring in 
an atom $\mathtt{A}$ over $\mathtt{r}/m$, and let $\sigma=\{\mathtt{V}_1/\mathtt{s}_1,\ldots \mathtt{V}_l/\mathtt{s}_l\}$ 
be a substitution grounding $\mathtt{A}$. Each $\mathtt{V}_{j}$ is then considered a random variable over the domain 
$D_{\arg(\mathtt{V}_j)}^{\mathtt{r}/m}$
of the argument $\arg(\mathtt{V}_j)$ it appears first in. 
Then, $\mu(\mathtt{A}\sigma\mid \mathtt{A})=\prod_{j=1}^{l}P_{\arg(\mathtt{V}_j)}^{\mathtt{r}/m}(\mathtt{s}_{j})$.
E.g.\, 
$\mu(\mathtt{emacs(hmm1,tex)}\mid \mathtt{emacs(F,E)})$,
is computed as the product of $P_{1}^{\mathtt{emacs}/2}(\mathtt{hmm1})$ and $P_{2}^{\mathtt{emacs}/2}(\mathtt{tex})$.

Thus far the semantics of a single abstract transition has been defined.
A LOHMM usually consists of multiple abstract transitions
and this creates a further complication.  
\begin{exam}
Consider 
${0.8: \mathtt{latex(File,tex) \xleftarrow{\tiny emacs(File)} 
emacs(File,tex)}}$ and ${0.4: \mathtt{dvi(File) \xleftarrow{\tiny emacs(File)} 
emacs(File,User)}}$. These two abstract transitions make conflicting statements 
about the state resulting from $\mathtt{emacs(hmm1,tex)}$. Indeed, according to the first
transition, the probability is $0.8$ that the resulting state is $\mathtt{latex(hmm1,tex)}$ and
according to the second one it assigns $0.4$ to $\mathtt{xdvi(hmm1)}$.
\end{exam}
There are essentially two ways to deal with this situation. On the one hand,
one might want to combine and normalize the two transitions and assign a
probability of~$\frac{2}{3}$ respectively~$\frac{1}{3}$\;. 
On the other hand, one might want to have only one rule firing. In this
paper, we chose the latter option because it allows us to
consider transitions more independently, it simplifies learning, and it yields locally interpretable models.
We employ the subsumption (or generality) relation
among the $\mathtt{B}$-parts of the two abstract transitions. 
Indeed, the $\mathtt{B}$-part of 
the first transition  $\mathtt{B}_1 = \mathtt{emacs(File,tex)}$ is more specific than that of the 
second transition $\mathtt{B}_2 = \mathtt{emacs(File,User)}$ because there exists a substitution 
$\theta = \{\mathtt{User}/\mathtt{tex}\}$ 
such that $\mathtt{B}_2\theta = \mathtt{B}_1$, i.e., $\mathtt{B}_2$ subsumes $\mathtt{B}_1$. 
Therefore $G_{\Sigma_1}(\mathtt{B}_1) \subseteq G_{\Sigma_1}(\mathtt{B}_2)$ and
the first transition can be regarded as more informative than the second one. It should therefore
be preferred over the second one when starting from $\mathtt{emacs(hmm1,tex)}$. 
We will also say that the first \emph{transition} is \emph{more specific} than the second one.
Remark that this \emph{generality} relation
imposes a partial order on the set of all transitions.
These considerations lead to the strategy of only considering the maximally specific
transitions that apply to a state in order to determine the successor states. 
This implements a kind of exception handling or default reasoning and 
is akin to \citeauthor{Katz:87}'s~\citeyear{Katz:87}
{\it back-off} $n$-gram models. In back-off $n$-gram models, the most detailed model that is deemed to provide 
sufficiently reliable information about the current context is used. That is, if one encounters an $n$-gram 
that is not sufficiently reliable, then back-off to use an $(n-1)$-gram; if that is not reliable either then back-off to level $n-2$,
etc.

The conflict resolution strategy will work properly
provided that the bodies of all maximally specific transitions (matching a given state)
represent the same abstract state. 
This can be enforced by requiring the \emph{generality} relation over the $\mathtt{B}$-parts to be 
closed under the {\it greatest lower bound} (glb) for each predicate, i.e., 
for each pair $\mathtt{B}_1,\mathtt{B}_2$ of bodies, such that $\theta=\mgu(\mathtt{B}_1,\mathtt{B}_2)$ exists, there is 
another body $\mathtt{B}$ (called lower bound) which subsumes $\mathtt{B}_1\theta$ (therefore also $\mathtt{B}_2\theta$) and 
is subsumed by $\mathtt{B}_1,\mathtt{B}_2$,
and if there is any other lower bound then it is subsumed by $\mathtt{B}$.
E.g., if the body of the second abstract transition in our example is 
$\mathtt{emacs(hmm1,User)}$ then the set of abstract transitions 
would not be closed under glb.

Finally, in order to specify a prior distribution over states, we assume
a finite set $\Upsilon$ of clauses of the form
$p:\mathtt{H}\leftarrow\mathtt{start}$
using a distinguished $\mathtt{start}$ symbol
such that $p$ is the probability of the LOHMM to start in a state of $G_{\Sigma}(\mathtt{H})$.

By now we are able to formally define {\it logical hidden Markov models}.
\begin{defn}\label{def:lohmm}
A {\it logical hidden Markov model} (LOHMM) is a tuple $(\Sigma,\mu,\Delta,\Upsilon)$ where 
$\Sigma$ is a logical alphabet, $\mu$ a selection probability over $\Sigma$, 
$\Delta$ is a set of abstract transitions, and $\Upsilon$ is
a set of abstract transitions encoding a prior distribution. Let  
$\mathbf{B}$ be the set of
all atoms that occur as body parts of transitions in $\Delta$.
We assume $\mathbf{B}$ to be closed under glb and require 
\begin{equation}
\forall \mathtt{B}\in \mathbf{B}: \sum\nolimits_{p:\mathtt{H\xleftarrow{O}B}\in\Delta}p=1.0
\end{equation} 
and that the probabilities $p$ of clauses in $\Upsilon$ sum up to $1.0$\;.
\end{defn}  
HMMs are a special cases of LOHMMs in which
$\Sigma$ contains only relation symbols of arity zero and the selection probability
is irrelevant. Thus, LOHMMs directly generalize HMMs. 

\begin{figure}[t]
\psfrag{START}[cc][cc]{\tiny $\mathtt{start}$}
\psfrag{EMACS1}[cc][cc]{\tiny $\mathtt{emacs(F,U)}$}
\psfrag{EMACS2}[cc][cc]{\tiny $\mathtt{emacs(F^\prime,U)}$} 
\psfrag{EMACS3}[cc][cc]{\tiny $\mathtt{emacs(F,tex)}$}
\psfrag{LATEX}[cc][cc]{\tiny $\mathtt{latex(F,tex)}$}
\psfrag{LS}[cc][cc]{\tiny $\mathtt{ls(U^\prime)}$}

\psfrag{EMACS1O1}[cc][cc]{\tiny $\mathtt{emacs(F)}:0.7$}
\psfrag{EMACS1O2}[cc][cc]{\tiny $\mathtt{emacs(F)}:0.3$}
\psfrag{EMACS3O1}[cc][cc]{\tiny $\mathtt{emacs(F)}:0.3$}
\psfrag{EMACS3O2}[cc][cc]{\tiny $\mathtt{emacs(F)}:0.1$}
\psfrag{EMACS3O3}[cc][cc]{\tiny $\mathtt{emacs(F)}:0.6$}
\psfrag{LATEXO1}[cc][cc]{\tiny $\mathtt{latex(F)}:0.6$}
\psfrag{LATEXO2}[cc][cc]{\tiny $\mathtt{latex(F)}:0.2$}
\psfrag{LATEXO3}[cc][cc]{\tiny $\mathtt{latex(F)}:0.2$}
\psfrag{LSO1}[cc][cc]{\tiny $\mathtt{ls}:0.6$}
\psfrag{LSO2}[cc][cc]{\tiny $\mathtt{ls}:0.4$}
\psfrag{STARTO1}[cc][cc]{\tiny $0.55$}
\psfrag{STARTO2}[cc][cc]{\tiny $0.45$}

\begin{center}
\setlength{\epsfxsize}{8.2cm}
\centerline{\epsfbox{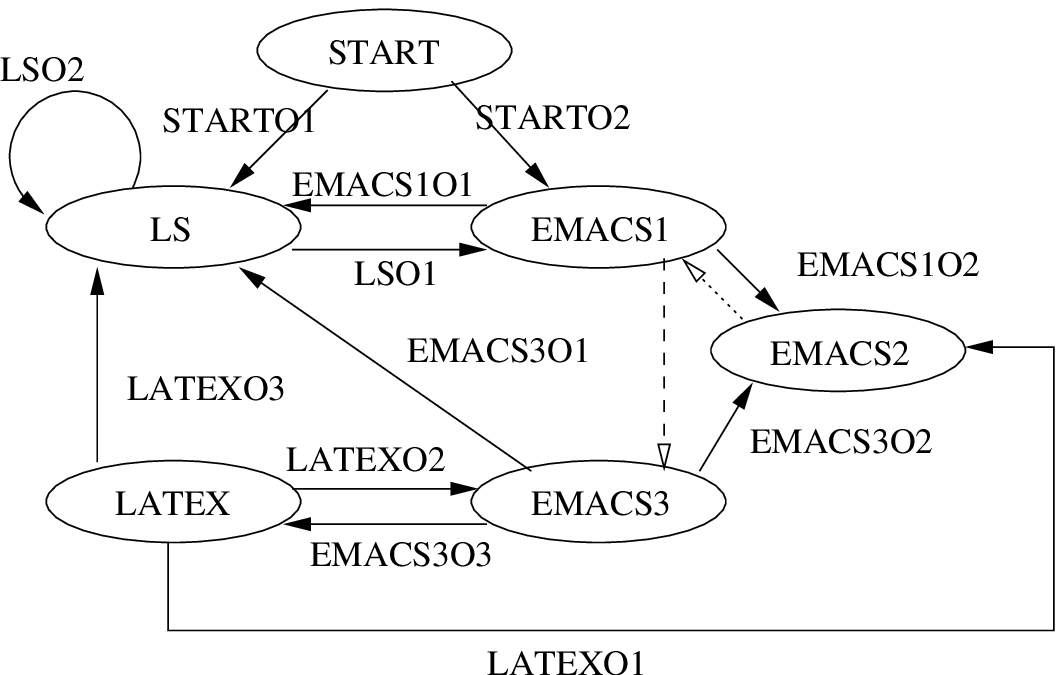}}
\caption{A logical hidden Markov model.\label{lhmm1}}
\end{center}
\end{figure}
LOHMMs can also be represented graphically. Figure~\ref{lhmm1} contains
an example. 
The underlying language $\Sigma_2$ consists of $\Sigma_1$ together
with the constant symbol $\mathtt{other}$ which denotes a user that does not employ \LaTeX. 
In this graphical notation, 
nodes represent abstract states and \emph{black tipped arrows} denote abstract transitions.
\emph{White tipped arrows} are used to represent meta knowledge.
More precisely, \emph{white tipped, dashed arrows} represent the generality or subsumption
ordering between abstract states. 
If we follow a transition to an abstract state with an outgoing \emph{white tipped, dotted arrow}
then this dotted arrow will always be followed. Dotted arrows are needed because the same
abstract state can occur under different circumstances. Consider the transition  
${p: \mathtt{latex(File^\prime,User^\prime) \xleftarrow{\tiny latex(File)} latex(File,User)}}$. Even though the atoms 
in the head and body of the transition are syntactically different they represent the same
abstract state. 
To accurately represent the meaning of this transition we cannot use
a black tipped arrow from 
\begin{figure}[t]
\psfrag{emacsFU}[cc][cc]{\scriptsize ${\mathtt{em(F,U)}}$}
\psfrag{emacslt}[cc][cc]{\scriptsize ${\mathtt{em(f_1,t)}}$}
\psfrag{emacsFt}[cc][cc]{\scriptsize ${\mathtt{em(F,t)}}$}
\psfrag{emacsl}[cc][cc]{\scriptsize ${\mathtt{em(f_1)}}$}
\psfrag{latexFt}[cc][cc]{\scriptsize ${\mathtt{la(F,t)}}$}
\psfrag{latexlt}[cc][cc]{\scriptsize ${\mathtt{la(f_1,t)}}$}
\psfrag{latexl}[cc][cc]{\scriptsize ${\mathtt{la(f_1)}}$}
\psfrag{emacsFpU}[cc][cc]{\scriptsize ${\mathtt{em(F^\prime,U)}}$}
\psfrag{emacsoo}[cc][cc]{\scriptsize ${\mathtt{em(f_2,o)}}$}
\psfrag{emacso}[cc][cc]{\scriptsize ${\mathtt{em(f_2)}}$}
\psfrag{lsUp}[cc][cc]{\scriptsize ${\mathtt{ls(U^\prime)}}$}
\psfrag{lst}[cc][cc]{\scriptsize ${\mathtt{ls(t)}}$}
\psfrag{ls}[cc][cc]{\scriptsize ${\mathtt{ls}}$}
\psfrag{mu}[cc][cc]{\scriptsize ${\mu}$}
\psfrag{mup}[cc][cc]{\scriptsize ${\mu}$}
\psfrag{mupp}[cc][cc]{\scriptsize ${\mu}$}
\psfrag{start}[cc][cc]{\scriptsize ${\mathtt{start}}$}
\psfrag{atom}[cc][cc]{\tiny abstract state}
\psfrag{state}[cc][cc]{\tiny state}
\begin{center}
\setlength{\epsfxsize}{10cm}
\centerline{\epsfbox{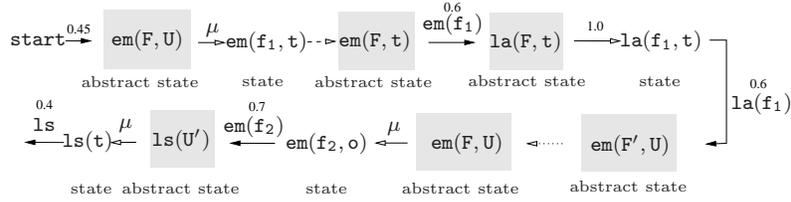}}
\caption{Generating the observation sequence $\mathtt{emacs(hmm1)},\mathtt{latex(hmm1)},$ $\mathtt{emacs(lohmm1)},\mathtt{ls}$ by the LOHMM in Figure~\ref{lhmm1}. The command $\mathtt{emacs}$ is abbreviated by $\mathtt{em}$, $\mathtt{f_1}$ denotes the filename $\mathtt{hmm1}$, $\mathtt{f_2}$ represents $\mathtt{lohmm1}$, $\mathtt{t}$ denotes a $\mathtt{tex}$ user, and $\mathtt{o}$ some $\mathtt{other}$ user. 
White tipped solid arrows indicate selections.\label{gen_path}}
\end{center}
\end{figure}
$\mathtt{latex(File,User)}$ to itself, because this 
would actually represent the abstract transition ${p:\mathtt{latex(File,User) \xleftarrow{\tiny latex(File)} latex(File,User)}}$.

Furthermore, the graphical representation clarifies that LOHMMs
are generative models. Let us explain how the model in Figure~\ref{lhmm1} would generate the
observation sequence 
$\mathtt{emacs(hmm1)},\mathtt{latex(hmm1)},$ $\mathtt{emacs(lohmm1)},\mathtt{ls}$
(cf. Figure~\ref{gen_path}).
It chooses an initial abstract state, say
$\mathtt{emacs(F,U)}$. Since both variables $\mathtt{F}$ and $\mathtt{U}$ are uninstantiated, the model
samples the state $\mathtt{emacs(hmm1,tex)}$ from $G_{\Sigma_2}$ using $\mu$. 
As indicated by the dashed arrow, $\mathtt{emacs(F,tex)}$ is more specific than $\mathtt{emacs(F,U)}$.
Moreover, $\mathtt{emacs(hmm1,tex)}$ matches $\mathtt{emacs(F,tex)}$. Thus, the model enters $\mathtt{emacs(F,tex)}$. 
Since the value of $\mathtt{F}$ was already instantiated in the previous abstract state, $\mathtt{emacs(hmm1,tex)}$ 
is sampled with probability $1.0$. 
Now, the model goes over to $\mathtt{latex(F,tex)}$, 
emitting $\mathtt{emacs(hmm1)}$ because the abstract observation
$\mathtt{emacs(F)}$ is already fully instantiated.
Again, since $F$ was already instantiated, $\mathtt{latex(hmm1,tex)}$
is sampled with probability $1.0$.
Next, we move on to $\mathtt{emacs(F^\prime,U)}$, 
emitting $\mathtt{latex(hmm1)}$. 
Variables $\mathtt{F^\prime}$ and $\mathtt{U}$ in $\mathtt{emacs(F^\prime,U)}$ 
were not yet bound; so, values, say $\mathtt{lohmm1}$ and $\mathtt{others}$, are sampled from $\mu$.
The dotted arrow brings us back to $\mathtt{emacs(F,U)}$. Because variables are implicitly universally 
quantified in abstract transitions, 
the scope of variables is restricted to single abstract transitions. In turn,  
$F$ is treated as a distinct, new variable,
and is automatically unified with $\mathtt{F^\prime}$, 
which is bound to $\mathtt{lohmm1}$. In contrast, variable $\mathtt{U}$ is already instantiated. 
Emitting $\mathtt{emacs(lohmm1)}$, the model makes a transition 
to $\mathtt{ls(U^\prime)}$. 
Assume that it samples $\mathtt{tex}$ for $\mathtt{U^\prime}$. Then, it remains
in $\mathtt{ls(U^\prime)}$ with probability $0.4$\;.
Considering all possible samples, 
allows one to prove the following theorem.
\begin{thm}[Semantics]\label{thm1}
A logical hidden Markov model over a language $\Sigma$ defines a discrete time 
stochastic process, i.e., a sequence of random variables $\langle X_t\rangle_{t=1,2,\ldots}$, where the domain of 
$X_t$ is $\hb(\Sigma)\times\hb(\Sigma)$. 
The induced probability measure over the Cartesian product $\bigotimes_t\hb(\Sigma)\times\hb(\Sigma)$ exists and is unique for each $t>0$ 
and in the limit~${t\rightarrow\infty}$.
\end{thm}

\noindent Before concluding this section, let us address some design choices underlying LOHMMs.

First, LOHMMs have been introduced as Mealy machines, i.e., output symbols are associated with transitions. 
Mealy machines fit our logical setting quite intuitively as they directly encode the conditional probability  
$P(\mathtt{O},\mathtt{S}^\prime |\mathtt{S})$ of making a transition from $\mathtt{S}$ to $\mathtt{S^\prime}$ 
emitting an observation $\mathtt{O}$. Logical hidden Markov models define this distribution as
\begin{equation*}
P(\mathtt{O},\mathtt{S}^\prime |\mathtt{S}) = \sum\nolimits_{p:\mathtt{H\xleftarrow{O^\prime}B}} p \ \cdot\ \mu(\mathtt{S^\prime}\mid \mathtt{H}\sigma_{\mathtt{B}}) \ \cdot \ \mu(\mathtt{O}\mid\mathtt{O^\prime}\sigma_{\mathtt{B}}\sigma_{\mathtt{H}})
\end{equation*}
where the sum runs over all abstract transitions $\mathtt{H\xleftarrow{O^\prime}B}$ such that
$\mathtt{B}$ is most specific for $\mathtt{S}$.
Observations correspond to (partially) observed proof steps and, hence, 
provide information shared among heads and bodies of abstract transitions.
In contrast, HMMs are usually introduced as {\it Moore} machines. Here, 
output symbols are associated with states implicitly assuming  $\mathtt{O}$ and $\mathtt{S^\prime}$ to be independent.
Thus, $P(\mathtt{O},\mathtt{S}^\prime\mid \mathtt{S})$ factorizes into $P(\mathtt{O}\mid \mathtt{S})\cdot P(\mathtt{S^\prime}\mid\mathtt{S})$. 
This makes it more difficult to observe information shared among heads and bodies. In turn,
Moore-LOHMMs are less intuitive and harder to understand.
For a more detailed discussion of the issue, we refer to Appendix~\ref{moore} where
we essentially show that  -- as in the propositional case -- Mealy- and Moore-LOHMMs are equivalent.

Second, the na\"\i ve Bayes approach for the selection distribution reduces the model complexity 
at the expense of a lower expressivity: functors are neglected and variables are treated independently.
Adapting more expressive approaches is an interesting future line of research. For instance, {\it Bayesian networks} allow one 
to represent {\it factorial HMMs}~\cite{GhahramaniJordan:97}. Factorial HMMs can be viewed as %a 
%special
%type of 
LOHMMs, where the hidden states are summarized by a $2\cdot k$-ary abstract state.
The first $k$ arguments encode the $k$ state variables, and
the last $k$ arguments serve as a memory of the previous joint state. 
$\mu$ of the $i$-th argument is conditioned on the $i+k$-th argument. 
{\it Markov chains} allow one to sample compound terms of  
finite depth such as $\mathtt{s(s(s(0)))}$ and 
to model e.g. misspelled filenames. This is akin to %the idea of 
{\it generalized HMMs}~\cite{KulpHausslerReeseEeckman:96}, in which each node 
may output a finite sequence of symbols rather than a single symbol.

Finally, LOHMMs --~as introduced in the present paper~-- specify a probability 
distribution over all sequences of a given length. Reconsider the LOHMM in Figure~\ref{lhmm1}.
Already the probabilities of all observation sequences of length $1$, i.e., 
$\mathtt{ls}$, $\mathtt{emacs(hmm1)}$, and $\mathtt{emacs(lohmm1)}$) sum up to $1$. More precisely, for each $t>0$
it holds that 
$\sum\nolimits_{x_1,\ldots,x_t}P(X_1=x_1,\ldots,X_t=x_t)=1.0\;$.
In order to model a distribution over sequences 
of variable length, i.e., 
$\sum\nolimits_{t>0}\sum\nolimits_{x_1,\ldots,x_t}P(X_1=x_1,\ldots,X_t=x_t)=1.0$
we may add a distinguished $\mathtt{end}$ state. The $\mathtt{end}$ state is absorbing in that
whenever the model makes a transition into this state, it terminates the observation sequence generated.

\section{Three Inference Problems for LOHMMs}\label{sec:inf}
\begin{figure}[t]
\psfrag{s0}[cc][cc]{\scriptsize $S_0$} 
\psfrag{s1}[cc][cc]{\scriptsize $S_1$} 
\psfrag{s2}[cc][cc]{\scriptsize $S_2$} 

\psfrag{start}[cc][cc]{\scriptsize $\mathtt{\scriptsize start}$}
\psfrag{sc1}[cc][cc]{\scriptsize $\mathtt{sc(1)}$} 
\psfrag{sc2}[cc][cc]{\scriptsize $\mathtt{sc(2)}$} 
\psfrag{scy}[cc][cc]{\scriptsize $\mathtt{sc(Y)}$} 
\psfrag{hc1}[cc][cc]{\scriptsize $\mathtt{hc(1)}$} 
\psfrag{hc2}[cc][cc]{\scriptsize $\mathtt{hc(2)}$} 
\psfrag{hcx}[cc][cc]{\scriptsize $\mathtt{hc(X)}$} 
\psfrag{scz}[cc][cc]{\scriptsize $\mathtt{sc(Z)}$} 
\psfrag{o1}[cc][cc]{\scriptsize $\mathtt{O}_1$} 
\psfrag{o2}[cc][cc]{\scriptsize $\mathtt{O}_2$} 
\psfrag{o3}[cc][cc]{\scriptsize $\mathtt{O}_2$} 
\psfrag{atom}[cc][cc]{\scriptsize abstract state}
\psfrag{state}[cc][cc]{\scriptsize states}
\begin{center}
\setlength{\epsfxsize}{8cm}
\centerline{\epsfbox{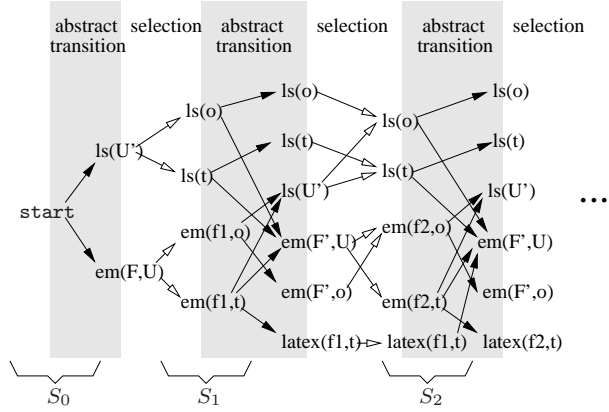}}
\caption{Trellis induced by the LOHMM in Figure~\ref{lhmm1}.
The sets of reachable states at time $0,1,\ldots $ are denoted by $S_0,S_1,\ldots$
In contrast with HMMs, there is an additional layer where the states are sampled from abstract states.
\label{fig:lattice4}}
\end{center}
\end{figure}

As for HMMs, three inference problems are of interest.
Let $M$ be a LOHMM and let $\mathsf{O}=\mathtt{O}_1,\mathtt{O}_2,\ldots,\mathtt{O}_T$, $T>0$, be a 
finite sequence of ground observations: 
\begin{description}
\item[(1) Evaluation:] Determine the 
probability $P(\mathsf{O}\mid M)$ that sequence $\mathsf{O}$ was generated by the model $M$.
\item[(2) Most likely state sequence:]
Determine the hidden state sequence $\mathsf{S}^*$ that has most likely produced the
observation sequence $\mathsf{O}$, i.e. 
${\mathsf{S}^*=\arg\max_{\mathsf{S}}{P(\mathsf{S}\mid\mathsf{O},M)}}\;$. 
\item[(3) Parameter estimation:] Given a set $\pmb{\mathsf{O}}=\{\mathsf{O}_1,\ldots,\mathsf{O}_k\}$ of observation sequences,
determine the most likely parameters $\lambda^*$ for the abstract transitions and the
selection distribution of $M$, i.e. 
$\lambda^*=\arg\max_{\lambda}{P(\pmb{\mathsf{O}}\mid \lambda)}\;$.
\end{description}
We will now address each of these problems in turn by upgrading the existing
solutions for HMMs. This will be realized by computing a grounded trellis as in Figure~\ref{fig:lattice4}.
The possible ground successor states of any given state are computed by 
first selecting the applicable abstract transitions and then applying the selection probabilities
(while taking into account the substitutions)
to ground the resulting states. This two-step factorization is coalesced into one step for HMMs.

To {\bf evaluate} $\mathsf{O}$, consider the probability 
of the partial observation sequence $\mathtt{O}_1,\mathtt{O}_2,\ldots,\mathtt{O}_{t}$ and
(ground) state $\mathtt{S}$ at time $t$, $0<t\leq T$, given the model $M=(\Sigma,\mu,\Delta,\Upsilon)$ %denoted by
\begin{equation*}
{\alpha_t(\mathtt{S}):={P(\mathtt{O}_1,\mathtt{O}_2,\ldots,\mathtt{O}_t,q_t=\mathtt{S}\mid M)}}
\end{equation*}
where $q_t=\mathtt{S}$ 
denotes that the system is in state $\mathtt{S}$ at time $t$. 
As for HMMs, $\alpha_t(\mathtt{S})$ can be computed using
a dynamic programming approach.
For $t=0$, we set 
${\alpha_0(\mathtt{S})={P(q_0=\mathtt{S}\mid M)}}\;$,
i.e., $\alpha_0(\mathtt{S})$ is the probability of starting in state~$\mathtt{S}$ and, for $t>0$,
we compute $\alpha_t(\mathtt{S})$ based on  $\alpha_{t-1}(\mathtt{S}^\prime)$:
\setlength{\fboxsep}{3pt}
\setlength{\fboxrule}{1.5pt}
\begin{tabbing} 
\hspace{4ex}\=\hspace{3ex}\=\hspace{3ex}\=\hspace{3ex}\=\hspace{3ex}\=\hspace{3ex}\=\hspace{3ex}\=\hspace{3ex}\=\hspace{3ex}  \kill
1:\>$S_0:=\{\mathtt{start}\}$ \> \> \> \> \> \> \> \hspace{1cm}{\em /* initialize the set of reachable states*/}\\
2:\> \keyw{for} $t=1,2,\ldots,T$ \keyw{do} \\
3:\> \vline \> $S_t=\emptyset$ \> \> \> \> \> \> \hspace{1cm}{\em /* initialize the set of reachable states at clock $t$*/}\\
4:\> \vline \> \keyw{foreach} $\mathtt{S}\in S_{t-1}$ \keyw{do} \\
5:\> \vline \> \vline \> \fbox{\keyw{foreach} maximally specific $p: \mathtt{H \xleftarrow{\tiny O} B} \in \Delta\cup\Upsilon$ s.t.
$\sigma_{\mathtt{B}} =\mgu(\mathtt{S},\mathtt{B})$ exists \keyw{do}} \\
6:\> \vline \> \vline \> \vline \> \fbox{\keyw{foreach} $\mathtt{S^\prime}=\mathtt{H}\sigma_{\mathtt{B}}\sigma_{\mathtt{H}} \in G_{\Sigma}(\mathtt{H}\sigma_{\mathtt{B}})$ s.t. $\mathtt{O}_{t-1}$ unifies with $\mathtt{O}\sigma_{\mathtt{B}}\sigma_{\mathtt{H}}$ \keyw{do}\hspace{6.7ex}} \\
7:\> \vline \> \vline \> \vline \> \vline \> \keyw{if} {$\mathtt{S^\prime}\not\in S_t$} \keyw{then}\\
8:\> \vline \> \vline \> \vline \> \vline \> \vline \> $S_t:=S_{t}\cup\{\mathtt{S^\prime}\}$\\
9:\> \vline \> \vline \> \vline \> \vline \> \vline \> $\alpha_{t}(\mathtt{S^\prime}):=0.0$ \\
10:\> \vline \> \vline \> \vline \> \vline \> $\alpha_{t}(\mathtt{S^\prime}) := \alpha_{t}(\mathtt{S^\prime}) \ + \ \alpha_{t-1}(\mathtt{S}) \ \cdot p \ \cdot\ $ \fbox{$\mu(\mathtt{S^\prime}\mid \mathtt{H}\sigma_{\mathtt{B}}) \ \cdot \ \mu(\mathtt{O}_{t-1}\mid \mathtt{O}\sigma_{\mathtt{B}}\sigma_{\mathtt{H}})$\hspace{3.7ex}}\\
11:\> \keyw{return} $P(\mathsf{O}\mid M)=\sum\nolimits_{\mathtt{S}\in S_{T}}\alpha_{T}(\mathtt{S})$
\end{tabbing}
where we assume for the sake of simplicity $\mathtt{O}\equiv\mathtt{start}$ for each abstract transition
$p:\mathtt{H}\leftarrow\mathtt{start}\in\Upsilon$. Furthermore,
the boxed parts specify all the differences to the HMM formula: unification and $\mu$ are taken into account.

Clearly, as for HMMs
$P(\mathsf{O}\mid M)=\sum\nolimits_{\mathtt{S}\in S_{T}}\alpha_{T}(\mathtt{S})$
holds.
The computational complexity of this {\it forward procedure} is ${\mathcal{O}(T\cdot s \cdot (  \left|\mathbf{B}\right| + o \cdot g))=\mathcal{O}(T\cdot s^2)}$ 
where $s=\max_{t=1,2,\ldots,T}\left| S_t\right|$\;, 
$o$ is the maximal number of outgoing abstract transitions
with regard to an abstract state, and $g$ is the maximal number of ground instances of
an abstract state. In a completely analogous manner, one can devise a {\em backward procedure} to compute
\begin{equation*}
{\beta_t(\mathtt{S})={P(\mathtt{O}_{t+1},\mathtt{O}_{t+2},\ldots,\mathtt{O}_T\mid q_t=\mathtt{S},M)}\;}.
\end{equation*}
This will be useful for solving Problem~{\bf (3)}.

Having a forward procedure, it is straightforward to adapt the Viterbi algorithm 
as a solution to Problem~{\bf (2)}, i.e., for computing the {\bf most likely state sequence}.
Let $\delta_t(\mathtt{S})$ denote the highest probability 
along a single path at time $t$ which
accounts for the first $t$ observations and ends in state $\mathtt{S}$, i.e.,
\begin{equation*}
\delta_t(\mathtt{S}) = \max_{\mathtt{S}_0,\mathtt{S}_1,\ldots,\mathtt{S}_{t-1}}P(\mathtt{S}_0,\mathtt{S}_1,\ldots,\mathtt{S}_{t-1},\mathtt{S}_{t}=\mathtt{S},O_1,\ldots,O_{t-1}|M)\;.
\end{equation*}
The procedure for finding the most likely state sequence basically follows the forward procedure.
Instead of summing over all ground transition probabilities in line 10, we maximize over them. 
More precisely, 
we proceed as follows:
\setlength{\fboxsep}{3pt}
\setlength{\fboxrule}{1.5pt}
\begin{tabbing} 
\hspace{4ex}\=\hspace{3ex}\=\hspace{3ex}\=\hspace{3ex}\=\hspace{3ex}\=\hspace{3ex}\=\hspace{3ex}\=\hspace{3ex}\=\hspace{3ex}  \kill
1:$S_0:=\{\mathtt{start}\}$  \> \> \> \> \> \> \> \hspace{1cm}{\em /* initialize the set of reachable states*/}\\
2:\> \keyw{for} $t=1,2,\ldots,T$ \keyw{do} \\
3:\> \vline \> $S_t=\emptyset$ \> \> \> \> \> \> \hspace{1cm}{\em /* initialize the set of reachable states at clock $t$*/}\\
4:\> \vline \> \keyw{foreach} $\mathtt{S}\in S_{t-1}$ \keyw{do} \\
5:\> \vline \> \vline \> \keyw{foreach} maximally specific $p: \mathtt{H \xleftarrow{\tiny O} B} \in \Delta\cup\Upsilon$ s.t.
$\sigma_{\mathtt{B}} =\mgu(\mathtt{S},\mathtt{B})$ exists \keyw{do} \\
6:\> \vline \> \vline \> \vline \> \keyw{foreach} $\mathtt{S^\prime}=\mathtt{H}\sigma_{\mathtt{B}}\sigma_{\mathtt{H}} \in G_{\Sigma}(\mathtt{H}\sigma_{\mathtt{B}})$ s.t. $\mathtt{O}_{t-1}$ unifies with $\mathtt{O}\sigma_{\mathtt{B}}\sigma_{\mathtt{H}}$ \keyw{do} \\
7:\> \vline \> \vline \> \vline \> \vline \> \keyw{if} {$\mathtt{S^\prime}\not\in S_t$} \keyw{then}\\
8:\> \vline \> \vline \> \vline \> \vline \> \vline \> $S_t:=S_{t}\cup\{\mathtt{S^\prime}\}$\\
9:\> \vline \> \vline \> \vline \> \vline \> \vline \> $\delta_t(\mathtt{S},\mathtt{S^\prime}):=0.0$\\
10:\> \vline \> \vline \> \vline \> \vline \> $\delta_{t}(\mathtt{S},\mathtt{S^\prime}) := \delta_{t}(\mathtt{S},\mathtt{S^\prime})+  \delta_{t-1}(\mathtt{S}) \cdot p \ \cdot\ \mu(\mathtt{S^\prime}\mid \mathtt{H}\sigma_{\mathtt{B}}) \ \cdot \ \mu(\mathtt{O}_{t-1}\mid \mathtt{O}\sigma_{\mathtt{B}}\sigma_{\mathtt{H}})$\\
11:\> \vline \> \keyw{foreach}  $\mathtt{S^\prime}\in S_{t}$ \keyw{do} \\
12:\> \vline \> \vline \> $\delta_t(\mathtt{S^\prime})=\max_{\mathtt{S}\in S_{t-1}}\delta_{t}(\mathtt{S},\mathtt{S^\prime})$\\
13:\> \vline \> \vline \> $\psi_t(\mathtt{S^\prime})=\arg\max_{\mathtt{S}\in S_{t-1}}\psi_{t}(\mathtt{S},\mathtt{S^\prime})$
\end{tabbing}
Here, $\delta_{t}(\mathtt{S},\mathtt{S^\prime})$ stores the probability of making a transition from
$\mathtt{S}$ to $\mathtt{S^\prime}$ and
$\psi_t(\mathtt{S^\prime})$ (with $\psi_1(S)=\mathtt{start}$ for all states $S$) keeps track of the state maximizing 
the probability along a single path at time $t$ which
accounts for the first $t$ observations and ends in state~$\mathtt{S^\prime}$.
The most likely hidden state sequence $\mathsf{S}^*$ can now be computed as
\begin{eqnarray*}
\mathtt{S}_{T+1}^*&=&\arg\max_{\mathtt{S}\in S_{T+1}}\delta_{T+1}(\mathtt{S}) \\
\text{and }\mathtt{S}_t^*&=&\psi_t(\mathtt{S}_{t+1}^*)\text{ for $t=T,T-1,\ldots,1$}\;.
\end{eqnarray*}

One can also consider problem {\bf (2)} on a more abstract level.
Instead of considering all contributions of different
abstract transitions $\cl$ to a single ground transition from state $\mathtt{S}$ to state $\mathtt{S}^\prime$
in line 10, one
might also consider the most likely abstract transition only.  This is realized
by replacing line 10 in the forward procedure with 
\begin{equation*}
{\alpha_{t}(\mathtt{S^\prime}) := \max (\alpha_{t}(\mathtt{S^\prime}),\alpha_{t-1}(\mathtt{S}) \cdot p \cdot\ \mu(\mathtt{S^\prime}\mid \mathtt{H}\sigma_{\mathtt{B}})  \cdot  \mu(\mathtt{O}_{t-1}\mid \mathtt{O}\sigma_{\mathtt{B}}\sigma_{\mathtt{H}}))}\;.
\end{equation*}
This solves the problem of finding the {\bf ($\mathbf{2^\prime}$) most likely state and abstract transition sequence}:
\begin{quote}
Determine the sequence of states and abstract transitions  
$\mathsf{GT}^*=\mathtt\mathtt{S_0},\cl_0,\mathtt\mathtt{S_1},\cl_{1},\mathtt{S_2},\ldots,\mathtt{S_T},\cl_T,\mathtt{S_{T+1}}$
where there exists substitutions $\theta_i$ with $S_{i+1}\leftarrow S_{i}\equiv\cl_i\theta_i$ 
that has most likely produced the
observation sequence $\mathsf{O}$, i.e. 
${\mathsf{GT}^*=\arg\max_{\mathsf{GT}}{P(\mathsf{GT}\mid\mathsf{O},M)}}\;.$ 
\end{quote}
Thus, logical hidden Markov models also pose new types of inference problems.

For {\bf parameter estimation}, we have to estimate the maximum likelihood
transition probabilities and selection distributions. To estimate the former, we 
upgrade the well-known {\em Baum-Welch} algorithm~\cite{Baum:72} for estimating 
the maximum likelihood parameters of HMMs 
and probabilistic context-free grammars. 

For HMMs, the Baum-Welch algorithm 
computes the improved estimate $\overline{p}$ of the transition probability of some 
(ground) transition $\cl\equiv p:\mathtt{H \xleftarrow{\tiny O} B}$ by taking the
ratio 
\begin{equation}\label{bw}
\overline{p}=\frac{\xi(\cl)}{\sum_{\mathtt{H^\prime \xleftarrow{\tiny O^\prime} B}\in \Delta\cup\Upsilon}\xi(\cl^\prime)}
\end{equation}
between the expected number $\xi(\cl)$ of times of making the 
transitions $\cl$ at any time given the model $M$ and an observation sequence $\mathsf{O}$, and 
the total number of times a transitions is made from $\mathtt{B}$ at any time given $M$ and $\mathsf{O}$. 

Basically the same applies when $\cl$ is an abstract transition.
However, we have to be a little bit more careful because we have no direct access to $\xi(\cl)$.
Let $\xi_t(\gcl,\cl)$
be the probability of following the abstract transition $\cl$ via its ground instance 
$\gcl\equiv p:\mathtt{GH\xleftarrow{\tiny GO} GB} $ at time $t$, i.e.,
\begin{equation}
\xi_t(\gcl,\cl)=\frac{\alpha_{t}(\mathtt{GB})\cdot p \cdot\beta_{t+1}(\mathtt{GH})}{ P(\mathsf{O}\mid M)}\cdot\fbox{$\mu(\mathtt{GH}\mid\mathtt{H}\sigma_{\mathtt{B}})\cdot\mu(\mathtt{O}_{t-1}\mid \mathtt{O}\sigma_{\mathtt{B}}\sigma_{\mathtt{H}})$}\;, \label{eqn2}
\end{equation}
where $\sigma_{\mathtt{B}}, \sigma_{\mathtt{H}}$ are as in the forward procedure (see above) and
$P(\mathsf{O}\mid M)$ is the probability that the model generated the sequence $\mathsf{O}$.
Again, the boxed terms constitute the main difference to the corresponding HMM formula.
In order to apply Equation~\eqref{bw} to compute improved estimates of probabilities associated with abstract
transitions, we set
\begin{equation*}
\xi(\cl)=\sum_{t=1}^T\xi_t(\cl)=\sum_{t=1}^T\sum_{\gcl}\xi_t(\gcl,\cl)
\end{equation*} 
where the inner sum runs over all ground instances of $\cl$.

This leads to the following re-estimation method, where we assume that the sets $S_i$ of reachable states
are reused from the computations of the $\alpha$-~and~$\beta$-values:
\begin{tabbing}
\hspace{4ex}\=\hspace{3ex}\=\hspace{3ex}\=\hspace{3ex}\=\hspace{3ex}\=\hspace{3ex}\=\hspace{3ex}\=\hspace{3ex}\=\hspace{3ex}  \kill
1:\>{\em /* initialization of expected counts */}\\ 
2:\> \keyw{foreach} $\cl\in \Delta\cup\Upsilon$ \keyw{do} \\
3:\> \vline \> $\xi(\cl):=m$ \hspace{1ex}{\em /* or $0$ if not using pseudocounts */}\\ 
4:\> {\em /* compute expected counts */} \\
5:\> \keyw{for} $t=0,1,\ldots ,T$ \keyw{do}\\
6:\> \vline \> \keyw{foreach} $\mathtt{S}\in S_t$ \keyw{do} \\ 
7:\> \vline \> \vline \> \fbox{\keyw{foreach} max. specific $\cl\equiv p: \mathtt{H \xleftarrow{\tiny O} B} \in \Delta\cup\Upsilon$ s.t. 
$\sigma_{\mathtt{B}} =\mgu(\mathtt{S},\mathtt{B})$ exists \keyw{do}\hspace{6.7ex}} \\
8:\> \vline \> \vline \> \vline \>\fbox{\keyw{foreach} $\mathtt{S^\prime}=\mathtt{H}\sigma_{\mathtt{B}}\sigma_{\mathtt{H}}\in G_{\Sigma}(\mathtt{H}\sigma_{\mathtt{B}})$ s.t. $\mathtt{S^\prime}\in S_{t+1}$
$\wedge$ $\mgu(\mathtt{O}_t,\mathtt{O}\sigma_{\mathtt{B}}\sigma_{\mathtt{H}})$ exists \keyw{do}\hspace{2.1ex}} \\  
9:\> \vline \> \vline \> \vline \> \vline \> $\xi(\cl) := \xi(\cl) + \alpha_{t}(\mathtt{S})\cdot p \cdot\beta_{t+1}(\mathtt{S^\prime})\big/ P(\mathsf{O}\mid M)\cdot$
\fbox{$\mu(\mathtt{S^\prime}\mid \mathtt{H}\sigma_{\mathtt{B}})\cdot\mu(\mathtt{O}_{t-1}\mid \mathtt{O}\sigma_{\mathtt{B}}\sigma_{\mathtt{H}})$} %\\ 
\end{tabbing}  

Here, equation~\eqref{eqn2} can be found in line~$9$. In line $3$, we set pseudocounts as
small sample-size regularizers. Other methods to avoid a biased underestimate of probabilities 
and even zero probabilities such as $m$-estimates (see e.g., \citeauthor{Mitchell:97}, 1997) can be easily adapted.

To estimate the selection probabilities, recall that $\mu$ follows a na\"\i ve Bayes scheme. 
Therefore, the estimated probability for a domain element $d\in D$ for some domain $D$ 
is the ratio between the number of times $d$ is selected and the number of times any
$d^\prime\in D$ is selected. The procedure for computing the $\xi$-values can thus be reused.
 
Altogether, the Baum-Welch algorithm works as follows:
While not converged, (1) estimate the
abstract transition probabilities, and (2) the selection probabilities.
Since it is an instance of the EM algorithm, 
it increases the likelihood of the data with every update, and according 
to~\citeA{mclachlan97em}, 
it is guaranteed to reach a stationary point. All standard techniques  
to overcome limitations of EM algorithms are applicable. 
The computational complexity 
(per iteration) is 
$\mathcal{O}(k\cdot(\alpha + d))=\mathcal{O}(k\cdot T\cdot s^2+k\cdot d)$ 
where $k$ is the number of sequences, $\alpha$ is the complexity of computing the $\alpha$-values (see above), 
and $d$ is the sum over the sizes of domains associated to predicates. 
Recently, \citeA{Kersting05UAI} combined the Baum-Welch algorithm 
with structure search for 
model selection of logical hidden Markov models
using {\it inductive logic programming}~\cite{MuggletonDeRaedt:94} refinement operators. The refinement operators
account for different abstraction levels which have to be explored.

\section{Advantages of LOHMMs}\label{sec:bene}
In this section, we will investigate the benefits of LOHMMs:
{\bf (1)} LOHMMs are strictly more expressive than HMMs, and
{\bf (2)}, using abstraction, logical variables and unification can be beneficial.
More specifically, with {\bf (2)}, we will show that 
\begin{description}
\item[(B1)] LOHMMs can be --- by design --- smaller than their propositional instantiations, and
\item[(B2)] unification can yield better log-likelihood estimates.
\end{description}

\subsection{On the Expressivity of LOHMMs}\label{sec:expr}
Whereas HMMs specify probability distributions over regular languages, LOHMMs specify probability distributions
over more expressive languages.
\begin{thm}\label{theorem:pcfg}
For any (consistent) probabilistic context-free grammar (PCFG) $G$ for some language $\mathcal{L}$ there 
exists a LOHMM $M$ s.t. $P_G(w)=P_M(w)$ for all $w\in\mathcal{L}$.
\end{thm}
The proof (see Appendix~\ref{app:pcfg}) makes use of abstract states of unbounded 'depth'. More precisely, functors are used to implement a stack.
Without functors, LOHMMs cannot encode PCFGs and, because the Herbrand base is finite, it can be proven that there
always exists an equivalent HMM. 

Furthermore, if functors are allowed, LOHMMs are strictly more expressive than PCFGs.
They can specify probability distributions over some languages that are context-sensitive:
\begin{equation*}
\begin{array}{rrcl}
1.0:&\mathtt{stack(s(0),s(0))} & \mathtt{\leftarrow} & \mathtt{start}\\
0.8:&\mathtt{stack(s(X),s(X))} & \mathtt{\xleftarrow{\tiny a}} & \mathtt{ stack(X,X)}\\
0.2:&\mathtt{unstack(s(X),s(X))} & \mathtt{\xleftarrow{\tiny a}}& \mathtt{stack(X,X)}\\
1.0:&\mathtt{unstack(X,Y)} & \mathtt{\xleftarrow{\tiny b}} &\mathtt{unstack(s(X),Y)}\\
1.0:&\mathtt{unstack(s(0),Y)} & \mathtt{\xleftarrow{\tiny c}} &\mathtt{unstack(s(0),s(Y))}\\
1.0:&\mathtt{end} & \mathtt{\xleftarrow{\tiny end}} &\mathtt{unstack(s(0),s(0))}
\end{array}
\end{equation*}
The LOHMM defines a distribution over $\{a^nb^nc^n\mid n>0\}$.

Finally, the use of logical variables also enables one to deal with
\emph{identifiers}.
Identifiers are special types of constants that denote objects. Indeed,
recall the \textsc{UNIX} command sequence $\mathtt{emacs \ lohmms.tex}, \ \mathtt{ls},$ $\mathtt{latex \ lohmms.tex},\ldots$ 
from the introduction.
The filename $\mathtt{lohmms.tex}$
is an identifier. Usually, the specific identifiers do not matter
but rather the fact that the same object occurs multiple times in the
sequence. LOHMMs can easily deal with identifiers by setting the selection probability
$\mu$ to a constant for the arguments in which identifiers can occur. Unification
then takes care of the necessary variable bindings.

\subsection{Benefits of Abstraction through Variables and Unification}\label{sec:ben}
Reconsider the domain of \textsc{UNIX} command sequences.
\textsc{Unix} users oftenly reuse a newly created directory in subsequent commands such
as in $\mathtt{mkdir(vt100x)},$ $\mathtt{cd(vt100x)}$, $\mathtt{ls(vt100x)}$\;.
Unification should allow us to elegantly employ this information because
it allows us to specify that, after observing the created directory, the model 
makes a transition into a state where the newly created 
directory is used:
\begin{equation*}
p_1:\mathtt{cd(Dir,mkdir)}\leftarrow\mathtt{mkdir(Dir,com)}
\text{ \ \  and \ \  }
p_2:\mathtt{cd(\_,mkdir)}\leftarrow\mathtt{mkdir(Dir,com)}
\end{equation*}
If the first transition is followed, the $\mathtt{cd}$ command will move to the newly created directory;
if the second transition is followed, it is not specified which directory $\mathtt{cd}$ will move to.
Thus, the LOHMM captures the reuse of created directories as an argument of future commands. 
Moreover, the LOHMM encodes the simplest possible case to show the benefits of unification. 
At any time, the observation sequence uniquely determines the state
sequence,
and functors are not used. Therefore, we left out the abstract output symbols associated with
abstract transitions. In total, the LOHMM $U$, modelling the reuse of directories, consists of $542$ parameters only 
but still covers more than $451000$ (ground) states, see Appendix~\ref{app:UNIX} for the complete model.
The compression in the number of parameters supports {\bf (B1)}. 

To empirically investigate the benefits of unification, we compare $U$ with
the variant $N$ of $U$ where no variables are shared, i.e.,
no unification is used such that for instance the first transition above is not allowed, see Appendix~\ref{app:UNIX}. 
$N$ has $164$ parameters less than $U$.
We computed the following zero-one win function
\begin{equation*}
f( \mathsf{O} ) = \begin{cases}
1& \text{if } \big[\log P_U(\mathsf{O})- \log P_N(\mathsf{O})\big] > 0 \\
0& \text{otherwise}
\end{cases}
\end{equation*}
leave-one-out cross-validated
on \textsc{Unix} shell logs collected by Greenberg~\citeyear{Greenberg:88}. Overall, the 
data consists of $168$ users of four groups: computer scientists, nonprogrammers, novices and others. 
About $300000$ commands have been logged with an average of $110$ sessions per user. We present here results for a subset of the data.
We considered all computer scientist sessions in which at least a single $\mathtt{mkdir}$
command appears. These yield $283$ logical sequences over in total $3286$ ground atoms. 
The LOO win was ${81.63\%}$. Other LOO statistics are also in favor of $U$:
\begin{center}
\begin{tabular}{|c|c|c|c|c|} \cline{2-5}
\multicolumn{1}{c}{}&\multicolumn{2}{|c|}{training} &\multicolumn{2}{|c|}{test}\\ \cline{2-5} 
\multicolumn{1}{c|}{}& $\log P(\pmb{\mathsf{O}})$ &$\log \frac{P_U(\pmb{\mathsf{O}})}{P_N(\pmb{\mathsf{O}})}$ & $\log P(\mathsf{O})$ &$\log \frac{P_U(\mathsf{O})}{P_N(\mathsf{O})}$ \\ \hline
$U$& $-11361.0$ & \multirow{2}{*}{$1795.3$} & $-42.8$&\multirow{2}{*}{ $7.91$}\\
$N$& $-13157.0$ &           & $-50.7$& \\\hline
\end{tabular}
\end{center}
Thus, although $U$ has $164$ parameters more than $N$, it shows a better generalization performance. This result supports 
{\bf (B2)}.
A pattern often found in $U$ was~\footnote{The sum of probabilities is not the same ($0.15+0.08=0.23\neq0.25$) because of the 
use of pseudo counts
and because of the subliminal non-determinism (w.r.t. abstract states) in $U$, i.e., in case that the first transition fires,
the second one also fires.}
\begin{equation*}
0.15:\mathtt{cd(Dir,mkdir)}\leftarrow\mathtt{mkdir(Dir,com)}
\text{ \ \ and \ \ }
0.08:\mathtt{cd(\_,mkdir)}\leftarrow\mathtt{mkdir(Dir,com)}
\end{equation*}
favoring changing to the directory just made. This knowledge cannot be captured in $N$
\begin{equation*}
\begin{array}{rrcl}
	0.25:&\mathtt{cd(\_,mkdir)}&\leftarrow&\mathtt{mkdir(Dir,com).}
\end{array}
\end{equation*}
The results clearly show that abstraction through variables and unification can be beneficial for some applications,
i.e., {\bf (B1)} and {\bf (B2)} hold.

\section{Real World Applications}\label{sec:eval}
Our intentions here are to investigate whether LOHMMs can be applied to real world domains. More precisely,
we will investigate whether benefits {\bf(B1)} and {\bf (B2)} can also be exploited in real world application domains.
Additionally, we will investigate whether
\begin{description}
\item[(B3)] LOHMMs are competitive with ILP algorithms that can also utilize unification and abstraction through variables, and 
\item[(B4)] LOHMMs can handle tree-structured data similar to PCFGs.
\end{description}
To this aim, we conducted experiments on two bioinformatics application domains: 
protein fold recognition~\cite{KerstingRaikoKramerDeRaedt:03} and mRNA signal structure detection~\cite{HorvathWrobelBohnebeck:01}.
Both application domains are multiclass problems with five different classes each.

\subsection{Methodology}
In order to tackle the multiclass problem with LOHMMs, we followed a {\it plug-in estimate}
approach.
Let $\{c_1,c_2,\ldots,c_k\}$ be the set of possible classes. Given a finite set of training examples 
$\{(x_i,y_i)\}_{i=1}^n \subseteq \mathcal{X} \times \{c_1,c_2,\ldots,c_n\}$,
one tries to find $f:\mathcal{X} \to \{c_1,c_2,\ldots,c_k\}$ 
\begin{equation}\label{plug}
f( x ) = \arg\max_{c\in\{c_1,c_2,\ldots,c_k\}}P( x \mid M, \lambda_c^{*} ) \cdot P(c)\;.
\end{equation}
with low approximation error on the
training data as well as on unseen examples. In Equation~\eqref{plug}, $M$ denotes the model structure which
is the same for all classes, $\lambda_c^*$ denotes the maximum likelihood parameters of $M$
for class $c$ estimated on the training examples with $y_i=c$ only, and $P(c)$ is the prior class distribution. 

We implemented the Baum-Welch algorithm (with pseudocounts $m$, see line~$3$) for maximum likelihood parameter estimation
using the Prolog system Yap-$4.4.4$.  In all experiments, we set $m=1$ and let the Baum-Welch algorithm 
stop if the change in log-likelihood was less than $0.1$ from one iteration to the next.
The experiments were ran on a Pentium-IV $3.2$~GHz Linux machine.

\subsection{Protein Fold Recognition}
Protein fold recognition is concerned with how proteins fold in nature, i.e., 
their three-dimensional structures.
This is an important problem as the biological functions of proteins
depend on the way they fold.
A common approach 
is to use database searches to find proteins (of known fold)
similar to a newly discovered protein (of unknown fold).
To facilitate protein fold recognition, 
several expert-based classification schemes of proteins have been developed that 
group the current set of known protein structures according to the similarity of 
their folds. For instance, the {\it structural classification of 
proteins}~\cite{HubbardMurzinBrennerChotia:97} (SCOP) database hierarchically organizes proteins 
according to their structures and evolutionary origin.
From a machine learning
perspective, SCOP induces a classification problem: given a protein of unknown fold,
assign it to the best matching group of the classification scheme.
This {\it protein fold classification} problem has been  
investigated by~\citeA{Turcotte_et_al_01a} based on the inductive logic programming (ILP) system PROGOL and by~\shortciteA{KerstingRaikoKramerDeRaedt:03} based on LOHMMs. 

The secondary structure of protein domains\footnote{A domain can be viewed as a sub-section of a protein which appears 
in a number of distantly related proteins and which can fold independently of the rest of the protein.} can elegantly be represented as logical sequences.
For example, the secondary structure of the Ribosomal protein L4 is represented~as
\begin{equation*}
 \begin{aligned}
&\mathtt{st(null,2)},\mathtt{he(right,alpha,6)}, \mathtt{st(plus,2)},\mathtt{he(right,alpha,4)}, \mathtt{st(plus,2)},\\
&\mathtt{he(right,alpha,4)}, \mathtt{st(plus,3)},\mathtt{he(right,alpha,4)},\mathtt{st(plus,1)},
\mathtt{he(hright,alpha,6)}
\end{aligned}
\end{equation*}
Helices of a certain type, orientation and length
$\mathtt{he(\mathit{HelixType},\mathit{HelixOrientation},\mathit{Length})}$,
and strands of a certain orientation and length
$\mathtt{st(\me{StrandOrientation},\mathit{Length})}$
are atoms over logical predicates. 
The application of traditional HMMs to such sequences
requires one to either ignore the structure of helices and strands, which
results in a loss of information, or to take all possible combinations (of arguments
such as orientation and length) into account, which leads to a combinatorial
explosion in the number of
parameters

The results reported by~\shortciteA{KerstingRaikoKramerDeRaedt:03} indicate 
that LOHMMs are well-suited for protein fold classification: the number of
parameters of a LOHMM can by an order of magnitude be smaller than
the number of a corresponding HMM ($120$ versus approximately $62000$) and 
the generalization performance, a $74\%$ accuracy, is comparable 
to \shortciteauthor{Turcotte_et_al_01a}'s  \citeyear{Turcotte_et_al_01a} result based on the ILP system Progol,
a $75\%$ accuracy. \shortciteA{KerstingRaikoKramerDeRaedt:03}, however, do not cross-validate their
results nor investigate -- as it is common in bioinformatics -- the impact of primary sequence similarity on the classification accuracy. 
For instance, the two most commonly requested ASTRAL subsets are the subset of sequences with less than 
$95\%$ identity to each other ($95$~cut) and with less than $40\%$ identity to each other ($40$~cut).
Motivated by this, we conducted the following new experiments.

The data consists of logical sequences of the secondary structure of protein 
domains. As in the work of~\shortciteA{KerstingRaikoKramerDeRaedt:03}, 
the task is to predict one of the five most populated SCOP 
folds of alpha and beta proteins (a/b): TIM beta/alpha-barrel (fold $1$), NAD(P)-binding Rossmann-fold domains
(fold $2$), Ribosomal protein L4 (fold $23$), Cysteine hydrolase (fold $37$), and Phosphotyrosine protein phosphatases I-like
(fold $55$). The class of a/b proteins consists of proteins with mainly parallel beta
sheets (beta-alpha-beta units). The data have been extracted automatically from the 
ASTRAL dataset version~$1.65$~\cite{astral:04} for the $95$~cut and for the $40$~cut. 
As in the work of~\shortciteA{KerstingRaikoKramerDeRaedt:03}, we consider strands and helices only, i.e., coils and isolated
strands are discarded. For the $95$ cut, this yields $816$ logical sequences consisting of in total  $22210$ ground atoms.
The number of sequences in the classes
are listed as $293$, $151$, $87$, $195$, and $90$.
For the $40$ cut, this yields $523$ logical sequences consisting of in total $14986$ ground atoms.
The number of sequences in the classes
are listed as $182$, $100$, $66$, $122$, and $53$.

\begin{figure}
\begin{center} 
\input{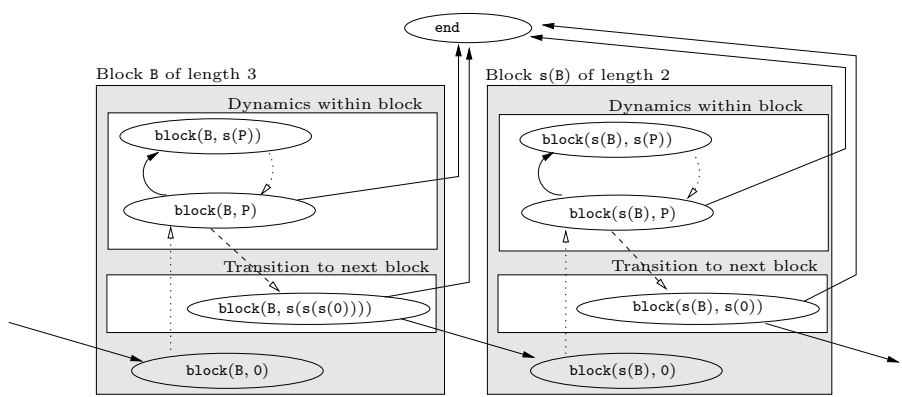} 
\end{center}
\caption{ Scheme of a left-to-right LOHMM block model.\label{fig:block}}
\end{figure}

\paragraph{LOHMM structure:}
The used LOHMM structure follows a left-to-right block topology, see Figure~\ref{fig:block}, to
model blocks of consecutive helices (resp. strands).
Being in a $\mathit{Block}$ of some size $s$, say $3$, the model will remain in the same block for $s=3$ time steps.  
A similar idea has been used to model haplotypes~\cite{KoivistoPerolaVariloHennahEkelundLukkPeltonenUkkonenMannila:02,KoivistoKiviojaMannilaRastasUkkonen:04}.
In contrast to common HMM block models~\cite{WonPruegelBennettKrogh:04}, the transition parameters are 
shared within each block and
one can ensure that the model makes a transition to the next state $\mathtt{s(\mathit{Block})}$ only at the end
of a block; in our example after exactly $3$ intra-block transitions.
Furthermore, there are specific abstract transitions for all helix types and strand
orientations to model the priori distribution, the intra- and the inter-block transitions.
The number of blocks and their sizes were chosen according to
the empirical distribution over sequence lengths in the data so that 
the beginning and the ending of protein domains was likely captured in detail. This yield the following 
block structure
\begin{center} 
\begin{picture}(0,0)%
\includegraphics{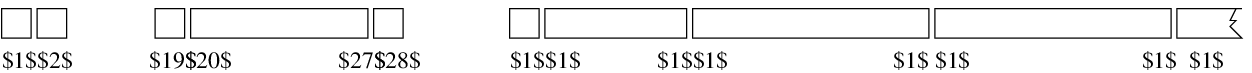}%
\end{picture}%
\setlength{\unitlength}{2818sp}%
\begingroup\makeatletter\ifx\SetFigFont\undefined%
\gdef\SetFigFont#1#2#3#4#5{%
  \reset@font\fontsize{#1}{#2pt}%
  \fontfamily{#3}\fontseries{#4}\fontshape{#5}%
  \selectfont}%
\fi\endgroup%
\begin{picture}(9474,457)(2644,-3206)
\put(2656,-3166){\makebox(0,0)[lb]{\smash{{\SetFigFont{8}{9.6}{\rmdefault}{\mddefault}{\updefault}{\color[rgb]{0,0,0}$1$}%
}}}}
\put(2926,-3166){\makebox(0,0)[lb]{\smash{{\SetFigFont{8}{9.6}{\rmdefault}{\mddefault}{\updefault}{\color[rgb]{0,0,0}$2$}%
}}}}
\put(6796,-3166){\makebox(0,0)[lb]{\smash{{\SetFigFont{8}{9.6}{\rmdefault}{\mddefault}{\updefault}{\color[rgb]{0,0,0}$41$}%
}}}}
\put(7651,-3166){\makebox(0,0)[lb]{\smash{{\SetFigFont{8}{9.6}{\rmdefault}{\mddefault}{\updefault}{\color[rgb]{0,0,0}$46$}%
}}}}
\put(7921,-3166){\makebox(0,0)[lb]{\smash{{\SetFigFont{8}{9.6}{\rmdefault}{\mddefault}{\updefault}{\color[rgb]{0,0,0}$47$}%
}}}}
\put(9451,-3166){\makebox(0,0)[lb]{\smash{{\SetFigFont{8}{9.6}{\rmdefault}{\mddefault}{\updefault}{\color[rgb]{0,0,0}$61$}%
}}}}
\put(9766,-3166){\makebox(0,0)[lb]{\smash{{\SetFigFont{8}{9.6}{\rmdefault}{\mddefault}{\updefault}{\color[rgb]{0,0,0}$62$}%
}}}}
\put(11341,-3166){\makebox(0,0)[lb]{\smash{{\SetFigFont{8}{9.6}{\rmdefault}{\mddefault}{\updefault}{\color[rgb]{0,0,0}$76$}%
}}}}
\put(11701,-3166){\makebox(0,0)[lb]{\smash{{\SetFigFont{8}{9.6}{\rmdefault}{\mddefault}{\updefault}{\color[rgb]{0,0,0}$77$}%
}}}}
\put(3781,-3166){\makebox(0,0)[lb]{\smash{{\SetFigFont{8}{9.6}{\rmdefault}{\mddefault}{\updefault}{\color[rgb]{0,0,0}$19$}%
}}}}
\put(4051,-3166){\makebox(0,0)[lb]{\smash{{\SetFigFont{8}{9.6}{\rmdefault}{\mddefault}{\updefault}{\color[rgb]{0,0,0}$20$}%
}}}}
\put(5221,-3166){\makebox(0,0)[lb]{\smash{{\SetFigFont{8}{9.6}{\rmdefault}{\mddefault}{\updefault}{\color[rgb]{0,0,0}$27$}%
}}}}
\put(5491,-3166){\makebox(0,0)[lb]{\smash{{\SetFigFont{8}{9.6}{\rmdefault}{\mddefault}{\updefault}{\color[rgb]{0,0,0}$28$}%
}}}}
\put(6481,-3166){\makebox(0,0)[lb]{\smash{{\SetFigFont{8}{9.6}{\rmdefault}{\mddefault}{\updefault}{\color[rgb]{0,0,0}$40$}%
}}}}
\put(3331,-2896){\makebox(0,0)[lb]{\smash{{\SetFigFont{20}{24.0}{\rmdefault}{\mddefault}{\updefault}{\color[rgb]{0,0,0}...}%
}}}}
\put(5941,-2896){\makebox(0,0)[lb]{\smash{{\SetFigFont{20}{24.0}{\rmdefault}{\mddefault}{\updefault}{\color[rgb]{0,0,0}...}%
}}}}
\end{picture}%
 
\end{center}
where the numbers denote the positions within protein domains. Furthermore, 
note that the last block gathers all remaining transitions. The blocks themselves are modelled using 
hidden abstract states over 
\begin{align*}
\mathtt{hc(\mathit{HelixType},\mathit{HelixOrientation},\mathit{Length},\mathit{Block})}
\text{ and } & \mathtt{sc(\mathit{StrandOrientation},\mathit{Length},\mathit{Block})}\;.
\end{align*}
Here, $\mathit{Length}$ denotes the number of consecutive bases the structure element consists of.
The length was discretized into $10$ bins such that the original lengths were uniformally distributed.
In total, the LOHMM has $295$ parameters. The corresponding HMM without parameter 
sharing has more than $65200$ parameters. This clearly confirms {\bf (B1)}.

\paragraph{Results:} We performed a $10$-fold cross-validation.
On the $95$ cut dataset, the accuracy was $76\%$ and took approx. $25$ minutes per cross-validation iteration;
on the $40$ cut, the accuracy was $73\%$ and took approx. $12$ minutes per cross-validation iteration.
The results validate \shortciteauthor{KerstingRaikoKramerDeRaedt:03}'s \citeyear{KerstingRaikoKramerDeRaedt:03} results and, in turn, 
clearly show that {\bf (B3)} holds.
Moreover, the novel results on the $40$ cut dataset indicate that
the similarities detected by the LOHMMs between the protein domain structures
were not accompanied by high sequence similarity.

\subsection{mRNA Signal Structure Detection}
\begin{figure}[t]
\psfrag{1.0}[rr][rr]{\tiny $\mathtt{helical(stem,n(n(n(n(n(n(n(0))))))),a,u).}$}
\psfrag{1.1}[rr][rr]{\tiny $\mathtt{helical(stem,n(n(n(n(n(n(0)))))),u,a).}$}
\psfrag{1.2}[rr][rr]{\tiny $\mathtt{helical(stem,n(n(n(n(n(0))))),c,a).}$}
\psfrag{1.3}[rr][rr]{\tiny $\mathtt{helical(stem,n(n(n(n(0)))),u,a).}$}
\psfrag{1.4}[rr][rr]{\tiny $\mathtt{helical(stem,n(n(n(0))),u,g).}$}
\psfrag{1.5}[rr][rr]{\tiny $\mathtt{helical(stem,n(n(0)),u,a).}$}
\psfrag{1.6}[rr][rr]{\tiny $\mathtt{helical(stem,n(0),a,a).}$}

\psfrag{2.0}[rr][rr]{\tiny $\mathtt{single(bulge5,n(0),a).}$}

\psfrag{3.0}[rr][rr]{\tiny $\mathtt{helical(stem,n(n(0)),c,g).}$}
\psfrag{3.1}[rr][rr]{\tiny $\mathtt{helical(stem,n(0),c,g).}$}

\psfrag{4.0}[rr][rr]{\tiny $\mathtt{single(bulge5,n(n(n(0))),g).}$}
\psfrag{4.1}[rr][rr]{\tiny $\mathtt{single(bulge5,n(n(0)),a).}$}
\psfrag{4.2}[rr][rr]{\tiny $\mathtt{single(bulge5,n(0),a).}$}

\psfrag{5.0}[rr][rr]{\tiny $\mathtt{helical(stem,n(n(n(0))),c,g).}$}
\psfrag{5.1}[rr][rr]{\tiny $\mathtt{helical(stem,n(n(0)),c,g).}$}
\psfrag{5.2}[rr][rr]{\tiny $\mathtt{helical(stem,n(0),c,g).}$}

\psfrag{6.0}[ll][ll]{\tiny $\mathtt{single(hairpin,n(n(n(0))),a).}$}
\psfrag{6.1}[ll][ll]{\tiny $\mathtt{single(hairpin,n(n(0)),u).}$}
\psfrag{6.2}[ll][ll]{\tiny $\mathtt{single(hairpin,n(0),u).}$}

\psfrag{7.0}[ll][ll]{\tiny $\mathtt{single(bulge3,n(0),a).}$}

\begin{center}
\begin{center}
\includegraphics{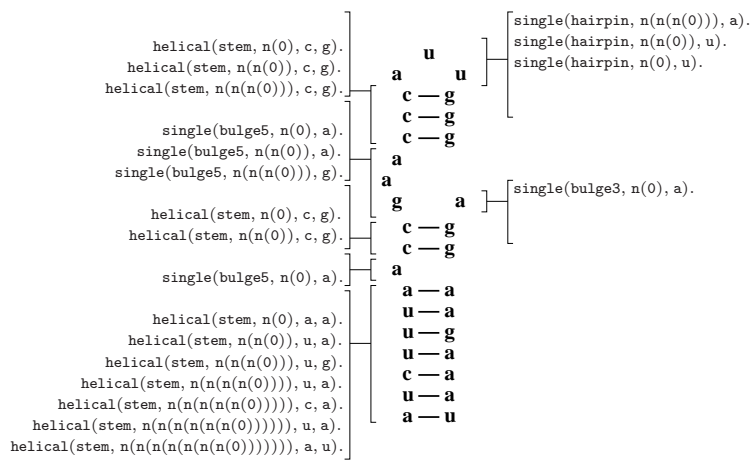}
\end{center}
\caption{The {\it chain} representation of a SECIS signal structure. The ground atoms are ordered clockwise 
starting with $\mathtt{helical(stem,n(n(n(n(n(n(n(0))))))),a,u)}$ at the lower left-hand side corner.\label{fig:mRNA_chain}}
\end{center}
\end{figure}

\begin{figure}[t]
\psfrag{0}[rr][rr]{\tiny $\mathtt{root(0,root,[c]).}$}

\psfrag{1.0}[rr][rr]{\tiny $\mathtt{helical(s(0),0,[c,c],stem,n(n(n(n(n(n(n(0)))))))).}$}
\psfrag{1.1}[rr][rr]{\tiny $\mathtt{nucleotide\_pair((a,u)).}$}
\psfrag{1.2}[rr][rr]{\tiny $\mathtt{nucleotide\_pair((u,a)).}$}
\psfrag{1.3}[rr][rr]{\tiny $\mathtt{nucleotide\_pair((c,a)).}$}
\psfrag{1.4}[rr][rr]{\tiny $\mathtt{nucleotide\_pair((u,a)).}$}
\psfrag{1.5}[rr][rr]{\tiny $\mathtt{nucleotide\_pair((u,g)).}$}
\psfrag{1.6}[rr][rr]{\tiny $\mathtt{nucleotide\_pair((u,a)).}$}
\psfrag{1.7}[rr][rr]{\tiny $\mathtt{nucleotide\_pair((a,a)).}$}

\psfrag{2.0}[rr][rr]{\tiny $\mathtt{single(s(s(0)),s(0),[],bulge5,n(0)).}$}
\psfrag{2.1}[rr][rr]{\tiny $\mathtt{nucleotide(a).}$}

\psfrag{3.0}[rr][rr]{\tiny $\mathtt{helical(s(s(s(0))),s(0),[c,c,c],stem,n(n(0))).}$}
\psfrag{3.1}[rr][rr]{\tiny $\mathtt{nucleotide\_pair((c,g)).}$}
\psfrag{3.2}[rr][rr]{\tiny $\mathtt{nucleotide\_pair((c,g)).}$}

\psfrag{4.0}[rr][rr]{\tiny $\mathtt{single(s(s(s(s(0)))),s(s(s(0))),[],bulge5,n(n(n(0)))).}$}
\psfrag{4.1}[rr][rr]{\tiny $\mathtt{nucleotide(g).}$}
\psfrag{4.2}[rr][rr]{\tiny $\mathtt{nucleotide(a).}$}
\psfrag{4.3}[rr][rr]{\tiny $\mathtt{nucleotide(a).}$}

\psfrag{5.0}[rr][rr]{\tiny $\mathtt{helical(s(s(s(s(s(0))))),s(s(s(0))),[c],stem,n(n(n(0)))).}$}
\psfrag{5.1}[rr][rr]{\tiny $\mathtt{nucleotide\_pair((c,g)).}$}
\psfrag{5.2}[rr][rr]{\tiny $\mathtt{nucleotide\_pair((c,g)).}$}
\psfrag{5.3}[rr][rr]{\tiny $\mathtt{nucleotide\_pair((c,g)).}$}

\psfrag{6.0}[ll][ll]{\tiny $\mathtt{single(s(s(s(s(s(s(0)))))),s(s(s(s(s(0))))),}$}
\psfrag{6.1}[ll][ll]{\tiny $\mathtt{[],hairpin,n(n(n(0)))).}$}
\psfrag{6.2}[ll][ll]{\tiny $\mathtt{nucleotide(a).}$}
\psfrag{6.3}[ll][ll]{\tiny $\mathtt{nucleotide(u).}$}
\psfrag{6.4}[ll][ll]{\tiny $\mathtt{nucleotide(u).}$}

\psfrag{7.0}[ll][ll]{\tiny $\mathtt{single(s(s(s(s(s(s(s(0))))))),s(s(s(0))),}$}
\psfrag{7.1}[ll][ll]{\tiny $\mathtt{[],bulge3,n(0)).}$}
\psfrag{7.2}[ll][ll]{\tiny $\mathtt{nucleotide(a).}$}

\psfrag{T0}[cc][cc]{\tiny $\mathtt{0}$}
\psfrag{T1}[cc][cc]{\tiny $\mathtt{s(0)}$}
\psfrag{T2}[cc][cc]{\tiny $\mathtt{s(s(0))}$}
\psfrag{T3}[cc][cc]{\tiny $\mathtt{s(s(s(0)))}$}
\psfrag{T4}[rr][rr]{\tiny $\mathtt{s(s(s(s(0))))}$}
\psfrag{T5}[cc][cc]{\tiny $\mathtt{s(s(s(s(s(0))))}$}
\psfrag{T6}[cc][cc]{\tiny $\mathtt{s(s(s(s(s(s(0))))))}$}
\psfrag{T7}[ll][ll]{\tiny $\mathtt{s(s(s(s(s(s(s(0)))))))}$}

\begin{center}
\begin{center}
\includegraphics{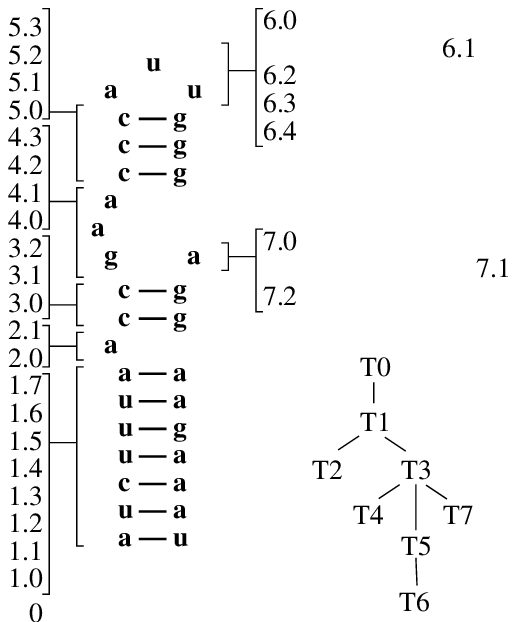}
\end{center}
\caption{ The {\it tree} representation of a SECIS signal structure. {\bf (a)} The logical sequence, i.e., the sequence of
ground atoms representing the SECIS signal structure. The ground atoms are ordered clockwise 
starting with $\mathtt{root(0,root,[c])}$ in the lower left-hand side corner. {\bf (b)} The tree
formed by the secondary structure elements.\label{fig:mRNA_tree}}
\end{center}
\end{figure}

mRNA sequences consist of bases ($\mathtt{ g}$uanine, $\mathtt{a}$denine, $\mathtt{u}$racil, 
$\mathtt{c}$ytosine) and fold intramolecularly to form a number of short base-paired 
stems~\cite{DurbinEddyKroghMitchison:98}. This base-paired structure is called the {\it secondary structure},
cf. Figures~\ref{fig:mRNA_chain} and~\ref{fig:mRNA_tree}.
The secondary structure contains special subsequences called signal structures that are responsible for special 
biological functions, such as RNA-protein interactions and cellular transport. The function of each signal structure
class is based on the common characteristic binding site of all class elements.
The elements are not necessarily identical but very similar. They can vary in 
topology (tree structure), in size (number of constituting bases), 
and in base sequence.

The goal of our experiments was to recognize instances of signal structures classes in
mRNA molecules. The first application of relational learning to
recognize the signal structure class of mRNA molecules
was described in the works of ~\citeA{BohnebeckHorvathWrobel:98} and of \shortciteA{HorvathWrobelBohnebeck:01}, 
where the relational instance-based
learner \textsc{RIBL} was applied. 
The dataset~\footnote{The dataset is not the
same as described in the work by \shortciteA{HorvathWrobelBohnebeck:01}
because we could not obtain the original dataset. We will compare to the smaller
data set used by~\shortciteauthor{HorvathWrobelBohnebeck:01}, which consisted of $66$ signal structures and is
very close to our data set. On a larger data set (with $400$ structures)~\shortciteauthor{HorvathWrobelBohnebeck:01}
report an error rate of $3.8\%\;$.} we used was similar to the one described by 
~\shortciteA{HorvathWrobelBohnebeck:01}.
It consisted of $93$ mRNA secondary structure sequences. More precisely, it was composed of $15$ and $5$ SECIS 
(Selenocysteine Insertion Sequence), 
$27$ IRE (Iron Responsive Element), 
$36$ TAR (Trans Activating Region) and $10$ histone stem loops constituting five classes.

The secondary structure is composed of different building blocks such as
stacking region, hairpin loops, interior loops etc. 
In contrast to the secondary structure of proteins that forms chains,
the secondary structure of mRNA forms a tree. As trees can not easily be handled using HMMs,
mRNA secondary structure data is more challenging than that of proteins.
Moreover, \shortciteA{HorvathWrobelBohnebeck:01}
report that making the tree structure available to RIBL as background knowledge had
an influence on the classification accuracy.
More precisely, using a simple chain representation RIBL achieved a $77.2\%$ leave-one-out cross-validation (LOO) accuracy 
whereas using the tree structure as background knowledge RIBL achieved a 
$95.4\%$ LOO accuracy. 

We followed \shortciteauthor{HorvathWrobelBohnebeck:01}'s experimental setup, that is, we adapted their
data representations
to LOHMMs and compared a chain model with a tree model. 

\paragraph{Chain Representation:}
In the {\it chain} representation (see also Figure~\ref{fig:mRNA_chain}), signal structures are described by
$\mathtt{single(\mathit{TypeSingle},\mathit{Position},\mathit{Acid})}$ or $\mathtt{helical(\mathit{TypeHelical},\mathit{Position},\mathit{Acid},\mathit{Acid})}$.
Depending on its type, a structure element is represented by either $\mathtt{single}/3$ or  $\mathtt{helical}/4$.
Their first argument $\mathit{TypeSingle}$ (resp. $\mathit{TypeHelical}$) specifies the type of the structure element,
i.e., $\mathtt{single}, \mathtt{bulge3},\mathtt{bulge5}, \mathtt{hairpin}$ (resp. $\mathtt{stem}$).
The argument $\mathit{Position}$
is the position of the sequence element within the corresponding structure element 
counted down, i.e.\footnote{$\mathtt{n}^m\mathtt{(0)}$ is shorthand
for the recursive application of the functor $\mathtt{n}$ on $\mathtt{0}$ $m$ times, i.e., for position~$m$.}, 
$\{\mathtt{n}^{13}\mathtt{(0)},\mathtt{n}^{12}\mathtt{(0)},\ldots,\mathtt{n}^{1}\mathtt{(0)}\}$.
The maximal position was set to $13$ as this was the maximal position observed in the data.
The last argument encodes the observed nucleotide (pair).

The used LOHMM structure 
follows again the left-to-right block
structure shown in Figure~\ref{fig:block}.
Its underlying idea is to model blocks of consecutive helical structure elements.
The hidden states are modelled using 
$\mathtt{single(\mathit{TypeSingle},\mathit{Position},\mathit{Acid},\mathit{Block})}$ and 
$\mathtt{helical(\mathit{TypeHelical},\mathit{Position},\mathit{Acid},\mathit{Acid},\mathit{Block})}$.
Being in a $\mathit{Block}$ of consecutive helical (resp. single) structure elements, 
the model will remain in the $\mathit{Block}$ or transition to a $\mathtt{single}$ element.
The transition to a single (resp. helical) element only occurs at $\mathit{Position}$ $\mathtt{n(0)}$.
At all other positions $\mathtt{n(\mathit{Position})}$, there were transitions from helical (resp. single) structure elements to
helical (resp. single) structure elements at $\mathit{Position}$ capturing the dynamics of the
nucleotide pairs (resp. nucleotides) within structure elements. 
For instance, the transitions for block $\mathtt{n(0)}$ at position $\mathtt{n(n(0))}$ were
\begin{equation*}
\begin{array}{lrcl}
a:& \mathtt{he(stem,n(0),X,Y,n(0))}&\xleftarrow{p_a:\mathtt{he(stem,n(0),X,Y)}}&\mathtt{ he(stem,n(n(0)),X,Y,n(0)))}\\
b:& \mathtt{he(stem,n(0),Y,X,n(0))}&\xleftarrow{p_b:\mathtt{he(stem,n(0),X,Y)}}&\mathtt{ he(stem,n(n(0)),X,Y,n(0)))}\\
c:& \mathtt{he(stem,n(0),X,\_,n(0))}&\xleftarrow{p_c:\mathtt{he(stem,n(0),X,Y)}}&\mathtt{ he(stem,n(n(0)),X,Y,n(0)))}\\
d:&  \mathtt{he(stem,n(0),\_,Y,n(0))}&\xleftarrow{p_d:\mathtt{he(stem,n(0),X,Y)}}&\mathtt{ he(stem,n(n(0)),X,Y,n(0)))}\\
e:&  \mathtt{he(stem,n(0),\_,\_,n(0))}&\xleftarrow{p_e:\mathtt{he(stem,n(0),X,Y)}}&\mathtt{ he(stem,n(n(0)),X,Y,n(0)))}
\end{array}
\end{equation*}
In total, there were $5$ possible blocks as this was the maximal number of blocks of consecutive helical
structure elements observed in the data. Overall, the LOHMM has $702$ parameters. In contrast, 
the corresponding HMM has more than $16600$ transitions validating {\bf (B1)}.

\paragraph{Results:} The LOO test log-likelihood was $-63.7$, 
%the LOO classification accuracy was $99\%$ ($92/93$), 
and an EM iteration took on average
$26$ seconds. 

Without the unification-based transitions $b$-$d$, i.e., 
using only the abstract transitions
\begin{equation*}
\begin{array}{lrcl}
a:& \mathtt{he(stem,n(0),X,Y,n(0))}&\xleftarrow{p_a:\mathtt{he(stem,n(0),X,Y)}}&\mathtt{ he(stem,n(n(0)),X,Y,n(0)))}\\
e:&  \mathtt{he(stem,n(0),\_,\_,n(0))}&\xleftarrow{p_e:\mathtt{he(stem,n(0),X,Y)}}&\mathtt{ he(stem,n(n(0)),X,Y,n(0)))},
\end{array}
\end{equation*}
the model has $506$ parameters. The LOO test log-likelihood was $-64.21$, %the LOO accuracy was $99\%$ ($92/93$), 
and an EM iteration took on average $20$ seconds.
The difference in LOO test log-likelihood is statistically significant (paired $t$-test, $p=0.01$).

Omitting even transition $a$, 
the  LOO test log-likelihood dropped to $-66.06$, %the LOO accuracy stayed at $99\%$ ($92/93$), 
and 
the average time per EM iteration was $18$ seconds. 
The model has $341$ parameters. The difference in average LOO log-likelihood is statistically significant 
(paired $t$-test, $p=0.001$).

The results clearly show that unification can yield better LOO test log-likelihoods, i.e., 
{\bf (B2)} holds.

\paragraph{Tree Representation:}

In the {\it tree} representation (see Figure~\ref{fig:mRNA_tree}~{\bf (a)}), the idea is to capture the tree structure
formed by the secondary structure elements, see Figure~\ref{fig:mRNA_tree}~{\bf (b)}. 
Each training instance is described as a sequence of ground facts over
\begin{align*}
&\mathtt{root(0,root,\mathit{\#Children})},\\
&\mathtt{helical(\mathit{ID},\mathit{ParentID},\mathit{\#Children},\mathit{Type},\mathit{Size})},\\
&\mathtt{nucleotide\_pair(\mathit{BasePair})},\\
&\mathtt{single(\mathit{ID},\mathit{ParentID},\mathit{\#Children},\mathit{Type},\mathit{Size})},\\
&\mathtt{nucleotide(\mathit{Base})}\;.
\end{align*}
Here,  $\mathit{ID}$ and $\mathit{ParentID}$ are natural numbers $\mathtt{0},\mathtt{s(0)},\mathtt{s(s(0))},\ldots$
encoding the child-parent relation, 
$\mathit{\#Children}$ denotes the number\footnote{Here, we use the Prolog short hand notation $\mathtt{[\cdot]}$ for lists. 
A list either is the constant $\mathtt{[]}$ representing the empty list, or is a compound term with functor $\mathtt{.}/2$ 
and two arguments, which are respectively the head and tail of the list. 
Thus $\mathtt{[a,b,c]}$ is the compound term
$\mathtt{.(a,.(b,.(c,[])))}$.}
of children $\mathtt{[]},\mathtt{[c]},
\mathtt{[c,c]},\ldots$, $\mathit{Type}$ is the type of the structure element such as $\mathtt{stem},\mathtt{hairpin},\ldots$,
and $\mathit{Size}$ is a natural number $\mathtt{0},\mathtt{n(0)},\mathtt{n(n(0))},\ldots$ Atoms %over
$\mathtt{root(0,root,\mathit{\#Children})}$ are used to root the topology.
The maximal $\mathit{\#Children}$
was $9$ and the maximal $\mathit{Size}$ was $13$ as this
was the maximal value observed in the data. 

As trees can not easily be handled using HMMs, we used a LOHMM which basically encodes a PCFG.
Due to Theorem~\ref{theorem:pcfg}, this is possible.
The used LOHMM structure can be found in Appendix~\ref{app:mRNA_tree}.
It processes the mRNA trees in in-order. Unification is only used for parsing the tree.
As for the chain representation, we used a $\mathit{Position}$ 
argument in the hidden states to encode the dynamics of nucleotides (nucleotide pairs) within secondary structure elements. 
The maximal $\mathit{Position}$ was again $13$.
In contrast to the chain representation, nucleotide pairs such as $\mathtt{(a,u)}$ are treated as constants. Thus,
the argument $\mathit{BasePair}$ consists of $16$ elements.

\paragraph{Results:} 
The LOO test log-likelihood was $-55.56$. 
Thus, exploiting the tree structure yields better probabilistic models. 
On average, an EM iteration took $14$ seconds. Overall, the result shows that {\bf (B4)} holds.\\

Although the Baum-Welch algorithm attempts to maximize a different objective function, namely the likelihood of
the data, it is interesting to compare LOHMMs and RIBL in terms of classification accuracy.
\paragraph{Classification Accuracy:}
On the {\it chain} representation, the LOO accuracies of all LOHMMs 
were $99\%$ ($92/93$). This is a considerable improvement on 
RIBL's $77.2\%$ ($51/66$) LOO accuracy for this representation.
On the {\it tree} representation, the LOHMM also achieved a LOO accuracy of $99\%$ ($92/93$). This  
is comparable to RIBL's LOO accuracy of $97\%$ ($64/66$) on this kind of representation.

Thus, already the chain LOHMMs show marked increases in LOO accuracy when compared to
RIBL~\shortcite{HorvathWrobelBohnebeck:01}. In order to achieve similar LOO accuracies,
\shortciteA{HorvathWrobelBohnebeck:01} had to make the tree structure available to RIBL as background knowledge. 
For LOHMMs, this had a significant influence on the LOO test log-likelihood, but not on the LOO accuracies.
This clearly supports {\bf (B3)}. Moreover, according to~\shortciteauthor{HorvathWrobelBohnebeck:01}, 
the mRNA application can also be considered a success in terms of the application domain, although
this was not the primary goal of our experiments. There exist also alternative parameter estimation techniques and other models, 
such as covariance models~\cite{EddyDurbin:94} or pair 
hidden Markov models~\cite{Sakakibara:03}, that might have been used as well as a basis for comparison.
However, as LOHMMs employ (inductive) logic programming principles, it is appropriate to compare with other
systems within this paradigm such as RIBL.

\section{Related Work}\label{sec:related}
LOHMMs combine two different research directions. On the one hand,
they are related to several extensions of HMMs and probabilistic grammars.
On the other hand, they are also related to 
the recent interest in combining inductive logic programming principles
with probability theory~\cite{DeRaedtKersting:03,DeRaedtKersting:04}.\\

In the first type of approaches, the underlying idea is to {\em upgrade} HMMs and probabilistic grammars
to represent more structured state spaces. 

Hierarchical HMMs~\cite{FineSingerTishby:98}, factorial HMMs~\shortcite{GhahramaniJordan:97}, and HMMs
based on tree automata~\cite{FrasconiSodaVullo:02} decompose the state variables into smaller units.
In hierarchical HMMs states themselves can be HMMs, in factorial
HMMs they can be factored into $k$ state variables which depend on one another
only through the observation, and in tree based HMMs the represented probability distributions 
are defined over tree structures. % consist of tree structures ???
The key difference with LOHMMs is that these approaches do not employ 
the logical concept of unification. Unification is essential because it
allows us to introduce abstract transitions, which do not {\em consist} of more detailed states.
As our experimental evidence shows, sharing information among abstract states
by means of unification can lead to more accurate model estimation.
The same holds for
{\em relational Markov models} (RMMs)~\cite{AndersonDomingosWeld:02}
to which LOHMMs are most closely related.
In RMMs, 
states can be of different types, with each type described by a different set of
variables. The domain of each variable
can be hierarchically structured. The main differences between LOHMMs and RMMs are that
RMMs do not either support variable binding nor unification nor hidden states. 

The equivalent of HMMs for context-free languages are {\it probabilistic context-free grammars} (PCFGs).
Like HMMs, they do not consider sequences of logical atoms and do not employ unification. Nevertheless,
there is a formal resemblance between the Baum-Welch algorithms for LOHMMs and 
for PCFGs. In case that a LOHMM encodes
a PCFG both algorithms are identical from a theoretical point of view. They
re-estimate the parameters 
as 
the ratio of the expected number of times a transition (resp. production) is used
and the expected number of times a transition (resp. production) might have been used.
The proof of Theorem~\ref{theorem:pcfg} assumes that the PCFG is given in Greibach normal form\footnote{A grammar is in GNF iff all productions are of the form $\mathtt{A}\leftarrow\mathtt{a}V$ where
$\mathtt{A}$ is a variable, $\mathtt{a}$ is exactly one terminal and $V$ is a string of none or more variables.} (GNF)
and uses a pushdown automaton to parse sentences. 
For grammars in GNF, pushdown automata are common for parsing.
In contrast, the actual computations of the Baum-Welch algorithm for PCFGs, the so called
Inside-Outside algorithm~\cite{Baker:79,LariYoung:90}, is usually formulated for grammars in 
{\it Chomsky normal form}\footnote{A grammar is in CNF iff every production
is of the form $\mathtt{A}\leftarrow\mathtt{B},\mathtt{C}$ or  $\mathtt{A}\leftarrow\mathtt{a}$  where
$\mathtt{A},\mathtt{B}$ and $\mathtt{C}$ are variables, and $\mathtt{a}$ is a terminal.}. The Inside-Outside
algorithm can make use of the efficient CYK algorithm~\cite{HopcroftUllman:79} for parsing strings.

An alternative to learning PCFGs from strings only is to learn from more structured data such as {\it skeletons}, which
are derivation trees with the nonterminal nodes removed~\cite{LevyJoshi:78}. Skeletons are exactly the
set of trees accepted by {\it skeletal tree automata} (STA). Informally, an STA, when given a tree as input, 
processes the tree bottom up, assigning a state to each node based on the states of that node's children. 
The STA accepts a tree iff it assigns a final state to the root of the tree. Due to this automata-based
characterization of the skeletons of derivation trees, the learning problem of (P)CFGs can be reduced to the 
problem of an STA. In particular, STA techniques have been adapted to learning tree 
grammars and (P)CFGs~\shortcite{Sakakibara:92,SakakibaraBrown:94} efficiently.

PCFGs have been extended in several ways. Most closely related to LOHMMs are
{\it unification-based grammars} which have been extensively studied in computational linguistics.
Examples are (stochastic) attribute-value grammars~\cite{Abney97d},
probabilistic feature grammars~\cite{Goodman:97}, 
head-driven phrase structure grammars~\cite{PollardSag:94}, and
lexical-functional grammars~\cite{Bresnan:01}.
For learning within such frameworks,
methods from undirected graphical models are used; see the work of~\citeA{johnson:03} for a description of some recent work.
The key difference to LOHMMs is that 
only nonterminals are replaced with structured, more complex entities. Thus, observation sequences of flat symbols and not of 
atoms are modelled. Goodman's {\it probabilistic feature grammars} are an exception. They treat terminals and nonterminals as
vectors of features. No abstraction is made, i.e., the feature vectors are ground instances, and no 
unification can be employed. Therefore, the number of parameters that needs to be estimated becomes  
easily very large, data sparsity 
is a serious problem. Goodman applied smoothing to overcome the problem.

\begin{figure}[t] 
\begin{center} 
\input{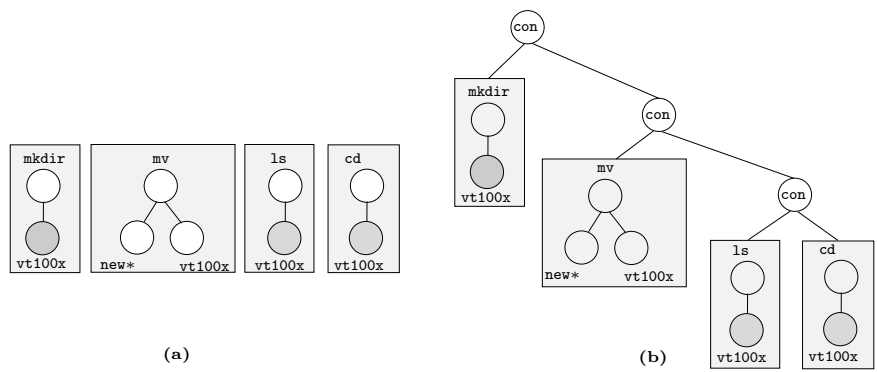}
\end{center}
\caption{{\bf (a)} Each atom in the logical sequence 
$\mathtt{mkdir(vt100x)}$, $\mathtt{mv(new*,vt100x)}$, $\mathtt{ls(vt100x)}$, $\mathtt{cd(vt100x)}$ 
forms a tree. The shaded nodes denote shared labels among the trees. {\bf (b)} The same sequence represented as a single tree. The predicate $\mathtt{con}/2$ represents the concatenation operator.\label{fig:tree_seq}}
\end{figure}

LOHMMs are generally related to (stochastic) tree automata (see e.g., \citeauthor{CarrascoOncinaCaleraRubio:01}, 2001).
Reconsider the \textsc{Unix} command sequence 
$\mathtt{mkdir(vt100x)},\mathtt{mv(new*,vt100x)},\mathtt{ls(vt100x)},\mathtt{cd(vt100x)}$\;. 
Each atom forms a tree, see Figure~\ref{fig:tree_seq}~{\bf (a)}, and, indeed, the whole sequence
of atoms also forms a (degenerated) tree, see Figure~\ref{fig:tree_seq}~{\bf (b)}.
Tree automata process single trees vertically, e.g.,  bottom-up. A state in the automaton is assigned to every node in the tree.
The state depends on the node label and on the states associated to the siblings of the node.
They do not focus on sequential domains. In contrast, LOHMMs are intended for learning in sequential domains.
They process sequences of trees horizontally, i.e., from left to right.
Furthermore, unification is used to share information between consecutive sequence elements. As Figure~\ref{fig:tree_seq}~{\bf (b)} 
illustrates, tree automata can only employ this information when allowing higher-order transitions, i.e., states depend on 
their node labels and on the states associated to predecessors $1,2,\ldots$ levels down the tree.\\

In the second type of approaches,
most attention has been devoted to developing highly expressive formalisms, such as e.g. PCUP~\cite{Eisele:94}, PCLP~\cite{Riezler:98},
SLPs~\cite{Muggleton:96}, 
PLPs~\cite{NgoHaddawy:97}, RBNs~\cite{Jaeger:97}, PRMs~\cite{FriedmanGetoorKollerPfeffer:99},
PRISM~\cite{Sato01parameterlearning}, BLPs~\cite{KerstingDeRaedt:01,KerstingDeRaedt:01b}, and DPRMs~\cite{SangheiDomingosWeld:03}.  
LOHMMs can be seen as an attempt towards {\em downgrading} such highly expressive frameworks.
Indeed, applying the main idea underlying LOHMMs to non-regular probabilistic grammar, i.e.,  
replacing flat symbols with atoms, yields -- in principle -- stochastic logic programs~\cite{Muggleton:96}.
As a consequence, LOHMMs represent
an interesting position on the expressiveness scale. Whereas they retain most of the essential
logical features of the more expressive formalisms, they seem easier to understand, adapt and learn.
This is akin to many contemporary considerations in 
{\em inductive logic programming}~\shortcite{MuggletonDeRaedt:94} and multi-relational data mining~\shortcite{DzeroskiLavrac:01}.

\section{Conclusions}\label{sec:future}
Logical hidden Markov models, a new formalism for representing
probability distributions over sequences of logical atoms, have been introduced
and solutions to the three central inference problems (evaluation, most likely state sequence
and parameter estimation) have been provided.
Experiments have demonstrated that unification can improve generalization accuracy, 
that the number of parameters of a LOHMM can be an order of magnitude smaller than the
number of parameters of the corresponding HMM, 
that the solutions presented
perform well in practice and also that LOHMMs possess several advantages over
traditional HMMs for applications involving structured sequences.

\paragraph*{Acknowledgments}
The authors thank Andreas Karwath and Johannes Horstmann for interesting
collaborations on the protein data; Ingo Thon for interesting collaboration on analyzing the \textsc{Unix} command sequences;
and Saul Greenberg for providing the \textsc{Unix} command sequence data.
The authors would also like to thank the anonymous reviewers for comments which considerably improved the paper.
This research was partly supported by the European Union
IST programme under contract numbers~IST-2001-33053 and~FP6-508861 (Application
of Probabilistic Inductive Logic Programming (APrIL)~I and~II).
Tapani Raiko was supported by a Marie Curie fellowship at DAISY, HPMT-CT-2001-00251.

\begin{appendix}

\section{Proof of Theorem~\ref{thm1}}
\begin{figure}[t]
\psfrag{Z1}[cc][cc]{\scriptsize $Z_1$}
\psfrag{Z2}[cc][cc]{\scriptsize $Z_2$} 
\psfrag{Z3}[cc][cc]{\scriptsize $Z_3$} 
\psfrag{E1}[cc][cc]{\scriptsize $U_1$} 
\psfrag{E2}[cc][cc]{\scriptsize $U_2$} 
\psfrag{E3}[cc][cc]{\scriptsize $U_3$} 
\begin{center}
\setlength{\epsfxsize}{5cm}
\centerline{\epsfbox{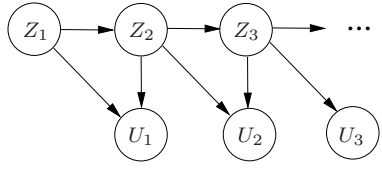}}
\caption{Discrete time stochastic process induced by a LOHMM. The nodes $Z_i$ and $U_i$ represent
random variables over $\hb_{\Sigma}$.\label{fig:dbn}}
\end{center}
\end{figure}

Let $M=(\Sigma,\mu,\Delta,\Upsilon)$ be a LOHMM.
To show that $M$ specifies a time discrete stochastic process, i.e., 
a sequence of random variables $\langle X_t\rangle_{t=1,2,\ldots}$, where the domains of 
the random variable $X_t$ is $\hb(\Sigma)$, the Herbrand base over $\Sigma$,
we define the {\it immediate state operator}
$T_{M}$-operator and the {\em current emission operator} $E_{M}$-operator. 
\begin{defn}{($T_{M}$-Operator, $E_{M}$-Operator )}
The operators $T_{M}:2^{\hb_{\Sigma}}\rightarrow 2^{\hb_{\Sigma}}$ and $E_{M}:2^{\hb_{\Sigma}}\rightarrow 2^{\hb_{\Sigma}}$ are
\begin{align*}
T_{M}(I)=\{&H\sigma_B\sigma_H\mid\exists (p:H \ \xleftarrow{O} \ B)\in M:
B\sigma_B\in I,H\sigma_B\sigma_H\in G_{\Sigma}(H)\}\\
E_{M}(I)=\{&O\sigma_B\sigma_H\sigma_O\mid\exists(p:H \ \xleftarrow{O} \ B)\in M:
B\sigma_B\in I, \ H\sigma_B\sigma_G\in G_{\Sigma}(H)\\&\text{ and }O\sigma_B\sigma_H\sigma_O\in G_{\Sigma}(O)\}
\end{align*}
\end{defn}
For each $i=1,2,3,\ldots$, the set 
$T^{i+1}_{M}(\{\mathtt{start}\}):=T_M(T_M^{i}(\{\mathtt{start}\}))$ with ${T^{1}_{M}(\{\mathtt{start}\}):=T_{M}(\{\mathtt{start}\})}$
specifies the state set at
clock $i$ which forms a random variable $Y_i$. The set $U^i_{M}(\{\mathtt{start}\})$ specifies the 
possible symbols emitted when transitioning from $i$ to $i+1$. It forms the variable $U_i$.
Each $Y_i$ (resp. $U_i$) can be extended to a random variable $Z_i$ (resp. $U_i$) over $\hb_{\Sigma}$:
\begin{equation*}
P(Z_i=z)=\left\{\begin{array}{rcl}0.0&:&z\not\in T^i_{M}(\{\mathtt{start}\})\\P(Y_i=z)&:& \text{otherwise}
\end{array}\right.
\end{equation*}
Figure~\ref{fig:dbn} depicts the influence relation among $Z_i$ and $U_i$.
Using standard arguments from probability theory and noting that
\begin{align*}
&P(U_{i}=U_i\mid Z_{i+1}=z_{i+1},Z_{i}=z_i)=\frac{P(Z_{i+1}=z_{i+1},U_{i}=u_{i}\mid Z_{i})}{\sum_{u_{i}} P(Z_{i+1},u_{i}\mid Z_{i})}\\
\text{ and }&P(Z_{i+1}\mid Z_{i}) =\sum_{u_{i}} P(Z_{i+1},u_{i}\mid Z_{i}) 
\end{align*}
where the probability distributions are due to equation~\eqref{trans_prob},
it is easy to show that Kolmogorov's
extension theorem (see~\citeauthor{Bauer:91}, 1991;~\citeauthor{FristedtGray:97}, 1997) holds. Thus, $M$ specifies a unique probability distribution over
$\bigotimes_{i=1}^t (Z_i\times U_i)$ for each $t>0$ 
and in the limit~${t\rightarrow\infty}$. $\hfill\Box$

\section{Moore Representations of LOHMMs}\label{moore}
For HMMs, {\it Moore} representations, i.e., output symbols are associated with states and
{\it Mealy} representations, i.e., output symbols are associated with transitions, are equivalent.
In this appendix, we will investigate to which extend this also holds for LOHMMs. 

Let $L$ be a Mealy-LOHMM according to definition~\ref{def:lohmm}. In the following, we will derive 
the notation of an equivalent LOHMM $L^\prime$ in Moore representation where there are
abstract transitions and abstract emissions (see below).
Each predicate $\mathtt{b}/n$ in $L$ is extended to $\mathtt{b}/n+1$ in $L^\prime$. The domains
of the first $n$ arguments are the same as for $\mathtt{b}/n$. The last argument will store
the observation to be emitted. More precisely, for each abstract transition 
\begin{equation*}
p:\mathtt{h(w_1,\ldots,w_l)}\xleftarrow{\mathtt{o(v_1,\ldots,v_k)}}\mathtt{b(u_1,\ldots,u_n)}
\end{equation*} 
in $L$, there is an abstract transition
\begin{equation*}
p:\mathtt{h(w_1,\ldots,w_l,o(v_1^\prime,\ldots,v_k^\prime))}\leftarrow \mathtt{b(u_1,\ldots,u_n,\_)}
\end{equation*} 
in $L^\prime$.
The primes in $\mathtt{o(v_1^\prime,\ldots,v_k^\prime)}$ denote that
we replaced each free~\footnote{A variable $\mathtt{X}\in\vars(\mathtt{o(v_1,\ldots,v_k)})$ is free iff
$\mathtt{X}\not\in\vars(\mathtt{h(w_1,\ldots,w_l)})\cup\vars(\mathtt{b(u_1,\ldots,u_n)})$.}
variables $\mathtt{o(v_1,\ldots,v_k)}$ by some distinguished constant symbol, say $\mathtt{\#}$.
Due to this, it holds that
\begin{equation}\label{eq:1}
\mu(\mathtt{h(w_1,\ldots,w_l)})=\mu(\mathtt{h(w_1,\ldots,w_l,o(v_1^\prime,\ldots,v_k^\prime))})\;,
\end{equation}
and $L^\prime$'s output distribution can be specified using {\it abstract emissions} which are expressions of the form
\begin{equation}\label{eq:3}
1.0:\mathtt{o(v_1,\ldots,v_k)}\leftarrow \mathtt{h(w_1,\ldots,w_l,o(v_1^\prime,\ldots,v_k^\prime))}\;.
\end{equation} 
The semantics of an abstract transition in $L^\prime$ is that being in some state 
${\mathtt{S^\prime_{\mathit t}} \in G_{\Sigma^\prime}(\mathtt{b(u_1,\ldots,u_n,\_)})}$ the system will make a transition into state 
${\mathtt{S^\prime_{\mathit{t+1}}} \in G_{\Sigma^\prime}(\mathtt{h(w_1,\ldots,w_l,o(v_1^\prime,\ldots,v_k^\prime))})}$
with probability 
\begin{equation}\label{eq:2}
p\cdot\mu(\mathtt{S^\prime_{\mathit{t+1}}}\mid \mathtt{h(w_1,\ldots,w_l,o(v_1^\prime,\ldots,v_k^\prime))}\mid\sigma_{\mathtt{S^\prime_{\mathit t}}})
\end{equation}
where ${\sigma_{\mathtt{S^\prime_{\mathit t}}}=\mgu(\mathtt{S^\prime_{\mathit t}},\mathtt{b(u_1,\ldots,u_n,\_)})}$. Due to Equation~\eqref{eq:1},
Equation~\eqref{eq:2} can be rewritten as 
\begin{equation*}
p\cdot\mu(\mathtt{S^\prime_{\mathit{t+1}}}\mid \mathtt{h(w_1,\ldots,w_l)}\mid\sigma_{\mathtt{S^\prime_{\mathit t}}})\;.
\end{equation*}
\noindent Due to equation~\eqref{eq:3}, the system will emit the output symbol 
${\mathtt{o_{\mathit{t+1}}}\in  G_{\Sigma^\prime}(\mathtt{o(v_1,\ldots,v_k)})}$ in state $\mathtt{S^\prime_{\mathit{t+1}}}$ with probability 
\begin{equation*}
\mu(\mathtt{o_{\mathit{t+1}}}\mid \mathtt{o(v_1,\ldots,v_k)}\sigma_{\mathtt{S^\prime_{\mathit{t+1}}}}\sigma_{\mathtt{S^\prime_{\mathit t}}})
\end{equation*} 
where ${\sigma_{\mathtt{S^\prime_{\mathit{t+1}}}}=\mgu(\mathtt{h(w_1,\ldots,w_l,o(v_1^\prime,\ldots,v_k^\prime))},\mathtt{S^\prime_{\mathit{t+1}}})}$.
Due to the construction of $L^\prime$, 
there exists a triple $(\mathtt{S_{\mathit t}},\mathtt{S_{\mathit{t+1}}},\mathtt{O_{t+1}})$ in $L$ 
for each triple $(\mathtt{S^\prime_{\mathit t}},\mathtt{S^\prime_{\mathit{t+1}}},\mathtt{O_{\mathit{t+1}}})$, $t>0$, in 
$L^\prime$ (and vise versa). Hence,both LOHMMs assign the same overall transition probability. 

$L$ and $L^\prime$ differ only in the way the
initialize sequences $\langle(\mathtt{S^\prime_{\mathit{t}}},\mathtt{S^\prime_{\mathit{t+1}}},\mathtt{O_{\mathit{t+1}}}\rangle_{t=0,2\ldots,T}$ 
(resp. $\langle(\mathtt{S_{\mathit{t}}},\mathtt{S_{\mathit{t+1}}},\mathtt{O_{t+1}}\rangle_{t=0,2\ldots,T}$). 
Whereas $L$ starts in some state $\mathtt{S_0}$ and makes a transition to $\mathtt{S_1}$ emitting $\mathtt{O_1}$,
the Moore-LOHMM $L^\prime$ is supposed to emit a symbol $\mathtt{O_0}$ in $\mathtt{S^\prime_0}$ before making 
a transition to $\mathtt{S^\prime_1}$. We compensate for this using the prior distribution.
The existence of the correct prior distribution for $L^\prime$ can be seen as follows. In $L$, there are only
finitely many states reachable at time $t=1$, i.e, $P_{L}(q_0=\mathtt{S})>0$ holds for only 
a finite set of ground states $\mathtt{S}$. The probability $P_{L}(q_0=\mathtt{s})$ can be computed 
similar to $\alpha_1(\mathtt{S})$. We set $t=1$ in line $6$, neglecting the condition on 
$\mathtt{O}_{t-1}$ in line $10$, and dropping 
$\mu(\mathtt{O}_{t-1}\mid \mathtt{O}\sigma_{\mathtt{B}}\sigma_{\mathtt{H}})$ from line $14$.
Completely listing all states $\mathtt{S}\in S_1$ together with $P_{L}(q_0=\mathtt{S})$, i.e., 
%\begin{equation*}
${P_{L}(q_0=\mathtt{S}):\mathtt{S}\leftarrow\mathtt{start}\;}$,
%\end{equation*}
constitutes the prior distribution of $L^\prime$.

The argumentation basically followed the approach to transform a Mealy machine into a Moore machine (see e.g., \citeauthor{HopcroftUllman:79}, 1979).
Furthermore, the mapping of a Moore-LOHMM --~as introduced in the present section~-- into a Mealy-LOHMM is straightforward.

\section{Proof of Theorem~\ref{theorem:pcfg}}\label{app:pcfg}
Let $T$ be a terminal alphabet and $N$ a nonterminal alphabet.
A {\it probabilistic context-free grammar} (PCFG) $G$ consists of a 
distinguished start symbol $S\in N$ plus a finite set of productions of the form
$p:X\rightarrow\alpha$, where $X\in N$, $\alpha\in(N\cup T)^*$ and $p\in[0,1]$.
For all $X\in N$, $\sum_{:X\rightarrow\alpha} p = 1$. A PCFG defines a stochastic 
process with sentential forms as states, and leftmost rewriting steps as transitions.
We denote a single rewriting operation of the grammar by a single arrow 
$\rightarrow$. If as a result of one ore more rewriting operations we are able to
rewrite $\beta\in(N\cup T)^*$ as a sequence $\gamma\in(N\cup T)^*$ of 
nonterminals and terminals, then we write $\beta\Rightarrow^*\gamma$. 
The probability of this rewriting is the product of all probability 
values associated to productions used in the derivation. We 
assume $G$ to be consistent, i.e., that the sum of all probabilities
of derivations $S\Rightarrow^*\beta$ such that $\beta\in T^*$ sum to $1.0$.

We can assume that the PCFG $G$ is in Greibach normal form. This follows 
from~\shortciteauthor{AbneyMcAllesterPereira:99}'s~\citeyear{AbneyMcAllesterPereira:99} Theorem~$6$ because $G$ is consistent.
Thus, every production $P\in G$ is of the form $p:X\rightarrow a Y_1\ldots Y_n$ for some $n\geq 0$.
In order to encode $G$ as a LOHMM $M$, we introduce
(1) for each non-terminal symbol $X$ in $G$ a constant symbol $\mathtt{nX}$
and (2) for each terminal symbol $t$ in $G$ a constant symbol $\mathtt{t}$. 
For each production $P\in G$, we include an abstract transition of the form 
$p:\mathtt{stack([nY_1,\ldots,nY_n|S])\xleftarrow{\tiny a} stack([nX|S])}$,
if $n>0$, and ${p:\mathtt{stack(S)\xleftarrow{\tiny a} stack([nX|S])}}$, if $n=0$.
Furthermore, we include ${1.0:\mathtt{stack([s])\leftarrow start}}$ and 
${1.0:\mathtt{end\xleftarrow{end} stack([])}}$. 
It is now straightforward to prove by induction that $M$ and $G$ are equivalent.$\hfill\Box$

\section{Logical Hidden Markov Model for \textsc{Unix} Command Sequences}\label{app:UNIX}
The LOHMMs described below model \textsc{Unix} command sequences
triggered by $\mathtt{mkdir}$. To this aim, we transformed the original Greenberg data into a sequence
of logical atoms over 
$\mathtt{com}, \mathtt{mkdir(Dir,LastCom)}, \mathtt{ls(Dir,LastCom)},  \mathtt{cd(Dir,Dir,LastCom)},$ 
$\mathtt{cp(Dir,Dir,LastCom)}$ and $\mathtt{mv(Dir,Dir,LastCom)}$.
The domain of $\mathtt{LastCom}$ was $\{\mathtt{start},\mathtt{com},\mathtt{mkdir}, \mathtt{ls},\mathtt{cd},$ $\mathtt{cp},$ 
$\mathtt{mv}\}$. The domain of $\mathtt{Dir}$ consisted of all
argument entries for $\mathtt{mkdir}, \mathtt{ls},\mathtt{cd},$ $\mathtt{cp},$ 
$\mathtt{mv}$ in the original dataset. Switches, pipes, etc. were neglected,
and paths were made absolute. This yields $212$ constants in the domain of $\mathtt{Dir}$.  
All original commands, which were not
$\mathtt{mkdir}, \mathtt{ls},\mathtt{cd},$ $\mathtt{cp},$ or $\mathtt{mv}$,
were represented as $\mathtt{com}$. If $\mathtt{mkdir}$ did not appear within $10$ time steps
before a command $C\in\{\mathtt{ls},\mathtt{cd},$ $\mathtt{cp},$$\mathtt{mv}\}$,
$C$ was represented as $\mathtt{com}$. Overall, this yields more than $451000$ ground states
that have to be covered by a Markov model.

The ``unification'' LOHMM $U$ basically implements a second order Markov model, i.e., the probability
of making a transition depends upon the current state and the previous state. It has $542$ parameters and the following structure:
\begin{gather*}
	\begin{array}{cc}
		\begin{array}{rcl}
			\mathtt{com}&\leftarrow&\mathtt{start.}\\
			\mathtt{mkdir(Dir,start)}&\leftarrow&\mathtt{start.}\\
		        & \\
		\end{array} & 
		\begin{array}{rcl}
			\mathtt{com}&\leftarrow&\mathtt{com.}\\
			\mathtt{mkdir(Dir,com)}&\leftarrow&\mathtt{com.}\\
			\mathtt{end}&\leftarrow&\mathtt{com.}
		\end{array}
	\end{array}
\end{gather*}
Furthermore, for each 
$C\in\{\mathtt{start},\mathtt{com}\}$ there are
\begin{gather*}	
	\begin{array}{cc}
		\begin{array}{rcl}
			\mathtt{mkdir(Dir,com)}&\leftarrow&\mathtt{mkdir(Dir,}C\mathtt{).}\\
			\mathtt{mkdir(\_,com)}&\leftarrow&\mathtt{mkdir(Dir,}C\mathtt{).}\\
			\mathtt{com}&\leftarrow&\mathtt{mkdir(Dir,}C\mathtt{).}\\
			\mathtt{end}&\leftarrow&\mathtt{mkdir(Dir,}C\mathtt{).}\\
			\mathtt{ls(Dir,mkdir)}&\leftarrow&\mathtt{mkdir(Dir,}C\mathtt{).}\\
			\mathtt{ls(\_,mkdir)}&\leftarrow&\mathtt{mkdir(Dir,}C\mathtt{).}\\
			\mathtt{cd(Dir,mkdir)}&\leftarrow&\mathtt{mkdir(Dir,}C\mathtt{).}
		\end{array}
		\begin{array}{rcl}
			\mathtt{cd(\_,mkdir)}&\leftarrow&\mathtt{mkdir(Dir,}C\mathtt{).}\\
			\mathtt{cp(\_,Dir,mkdir)}&\leftarrow&\mathtt{mkdir(Dir,}C\mathtt{).}\\
			\mathtt{cp(Dir,\_,mkdir)}&\leftarrow&\mathtt{mkdir(Dir,}C\mathtt{).}\\
			\mathtt{cp(\_,\_,mkdir)}&\leftarrow&\mathtt{mkdir(Dir,}C\mathtt{).}\\
			\mathtt{mv(\_,Dir,mkdir)}&\leftarrow&\mathtt{mkdir(Dir,}C\mathtt{).}\\
			\mathtt{mv(Dir,\_,mkdir)}&\leftarrow&\mathtt{mkdir(Dir,}C\mathtt{).}\\
			\mathtt{mv(\_,\_,mkdir)}&\leftarrow&\mathtt{mkdir(Dir,}C\mathtt{).}
		\end{array}
	\end{array}
\end{gather*}
together with for each 
$C\in\{\mathtt{mkdir},\mathtt{ls},\mathtt{cd},\mathtt{cp},\mathtt{mv}\}$
and for each $C_1\in\{\mathtt{cd},\mathtt{ls}\}$ (resp. ${C_2\in\{\mathtt{cp},\mathtt{mv}\}}$)
\begin{gather*}
	\begin{array}{cc}
		\begin{array}{rcl}
			\mathtt{mkdir(Dir,com)}&\leftarrow&C_1\mathtt{(Dir,}C\mathtt{).}\\
			\mathtt{mkdir(\_,com)}&\leftarrow&C_1\mathtt{(Dir,}C\mathtt{).}\\
			\mathtt{com}&\leftarrow&C_1\mathtt{(Dir,}C\mathtt{).}\\
			\mathtt{end}&\leftarrow&C_1\mathtt{(Dir,}C\mathtt{).}\\
			\mathtt{ls(Dir,}C_1\mathtt{)}&\leftarrow&C_1\mathtt{(Dir,}C\mathtt{).}\\
			\mathtt{ls(\_,}C_1\mathtt{)}&\leftarrow&C_1\mathtt{(Dir,}C\mathtt{).}\\
			\mathtt{cd(Dir,}C_1\mathtt{)}&\leftarrow&C_1\mathtt{(Dir,}C\mathtt{).}\\ 
			\mathtt{cd(\_,}C_1\mathtt{)}&\leftarrow&C_1\mathtt{(Dir,}C\mathtt{).}\\
			\mathtt{cp(\_,Dir,}C_1\mathtt{)}&\leftarrow&C_1\mathtt{(Dir,}C\mathtt{).}\\
			\mathtt{cp(Dir,\_,}C_1\mathtt{)}&\leftarrow&C_1\mathtt{(Dir,}C\mathtt{).}\\
			\mathtt{cp(\_,\_,}C_1\mathtt{)}&\leftarrow&C_1\mathtt{(Dir,}C\mathtt{).}\\
			\mathtt{mv(\_,Dir,}C_1\mathtt{)}&\leftarrow&C_1\mathtt{(Dir,}C\mathtt{).}\\
			\mathtt{mv(Dir,\_,}C_1\mathtt{)}&\leftarrow&C_1\mathtt{(Dir,}C\mathtt{).}\\
			\mathtt{mv(\_,\_,}C_1\mathtt{)}&\leftarrow&C_1\mathtt{(Dir,}C\mathtt{).}\\
			 & & 
		\end{array}
		\begin{array}{rcl}
			\mathtt{mkdir(\_,com)}&\leftarrow&C_2\mathtt{(From,To,}C\mathtt{).}\\
			\mathtt{com}&\leftarrow&C_2\mathtt{(From,To,}C\mathtt{).}\\
			\mathtt{end}&\leftarrow&C_2\mathtt{(From,To,}C\mathtt{).}\\
			\mathtt{ls(From,}C_2\mathtt{)}&\leftarrow&C_2\mathtt{(From,To,}C\mathtt{).}\\
			\mathtt{ls(To,}C_2\mathtt{)}&\leftarrow&C_2\mathtt{(From,To,}C\mathtt{).}\\
			\mathtt{ls(\_,}C_2\mathtt{)}&\leftarrow&C_2\mathtt{(From,To,}C\mathtt{).}\\
			\mathtt{cd(From,}C_2\mathtt{)}&\leftarrow&C_2\mathtt{(From,To,}C\mathtt{).}\\
			\mathtt{cd(To,}C_2\mathtt{)}&\leftarrow&C_2\mathtt{(From,To,}C\mathtt{).}\\
			\mathtt{cd(\_,}C_2\mathtt{)}&\leftarrow&C_2\mathtt{(From,To,}C\mathtt{).}\\
			\mathtt{cp(From,\_,}C_2\mathtt{)}&\leftarrow&C_2\mathtt{(From,To,}C\mathtt{).}\\
			\mathtt{cp(\_,To,}C_2\mathtt{)}&\leftarrow&C_2\mathtt{(From,To,}C\mathtt{).}\\
			\mathtt{cp(\_,\_,}C_2\mathtt{)}&\leftarrow&C_2\mathtt{(From,To,}C\mathtt{).}\\
			\mathtt{mv(From,\_,}C_2\mathtt{)}&\leftarrow&C_2\mathtt{(From,To,}C\mathtt{).}\\
			\mathtt{mv(\_,To,}C_2\mathtt{)}&\leftarrow&C_2\mathtt{(From,To,}C\mathtt{).}\\
			\mathtt{mv(\_,\_,}C_2\mathtt{)}&\leftarrow&C_2\mathtt{(From,To,}C\mathtt{).}
		\end{array}
	\end{array}
\end{gather*}
Because all states are fully observable, we omitted the output symbols associated with clauses, and,
for the sake of simplicity, we omitted associated probability values.

The ``no unification'' LOHMM $N$ is the variant of $U$ where no variables were shared such as
\begin{gather*}
	\begin{array}{rcl}	
		\mathtt{mkdir(\_,com)}&\leftarrow&\mathtt{cp(From,To,}C\mathtt{).}\\
		\mathtt{com}&\leftarrow&\mathtt{cp(From,To,}C\mathtt{).}\\
		\mathtt{end}&\leftarrow&\mathtt{cp(From,To,}C\mathtt{).}\\
& &
       \end{array}
	\begin{array}{rcl}
		\mathtt{ls(\_,cp)}&\leftarrow&\mathtt{cp(From,To,}C\mathtt{).}\\
		\mathtt{cd(\_,cp)}&\leftarrow&\mathtt{cp(From,To,}C\mathtt{).}\\
		\mathtt{cp(\_,\_,cp)}&\leftarrow&\mathtt{cp(From,To,}C\mathtt{).}\\
		\mathtt{mv(\_,\_,cp)}&\leftarrow&\mathtt{cp(From,To,}C\mathtt{).}
	\end{array}
\end{gather*}
Because only transitions are affected, $N$ has $164$ parameters less than $U$, i.e., $378$.

\section{Tree-based LOHMM for mRNA Sequences}\label{app:mRNA_tree}
The LOHMM processes the nodes of mRNA trees in in-order. The structure of the LOHMM is shown at the end of the section.
There are copies of the shaded parts.
Terms are abbreviated using their starting alphanumerical; $\mathtt{tr}$ stands for $\mathtt{tree}$,
$\mathtt{he}$ for $\mathtt{helical}$, $\mathtt{si}$ for $\mathtt{single}$, $\mathtt{nuc}$
for $\mathtt{nucleotide}$, and $\mathtt{nuc\_p}$ for $\mathtt{nucleotide\_pair}$.

The domain of $\mathit{\#Children}$ covers the maximal branching factor found in the data, i.e.,
$\{\mathtt{[c]},\mathtt{[c,c]},\ldots,\mathtt{[c,c,c,c,c,c,c,c,c]}\}$; the domain of $\mathit{Type}$ 
consists of all types occurring in the data, i.e.,
$\{\mathtt{stem},\mathtt{single}, \mathtt{bulge3}, \mathtt{bulge5}, \mathtt{hairpin}\}$; 
and for $\mathit{Size}$, the domain covers the maximal length of a secondary structure element in the data, i.e.,
the longest sequence of consecutive bases respectively base pairs constituting a secondary structure element. 
The length was encoded as $\{\mathtt{n}^1\mathtt{(0)},\mathtt{n}^2\mathtt{(0)},\ldots,\mathtt{n}^{13}\mathtt{(0)}\}$
where $\mathtt{n}^m\mathtt{(0)}$ denotes the recursive application of the functor $\mathtt{n}$ $m$ times.
For $\mathit{Base}$ and
$\mathit{BasePair}$, the domains were the $4$ bases respectively the $16$ base pairs.
In total, there are $491$ parameters.
\begin{figure}[t,b]
\begin{center}\scriptsize
\rotatebox{90}{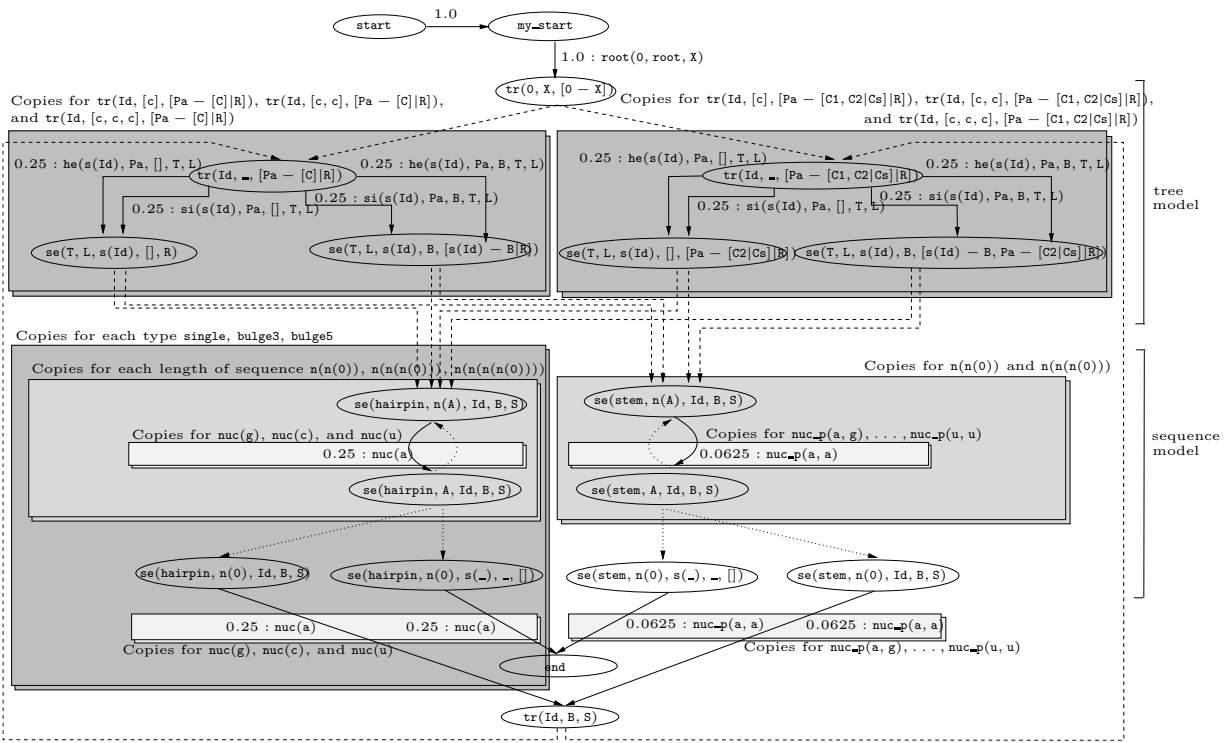}
\caption{The mRNA LOHMM structure. The symbol $\_$ denotes anonymous variables which are 
read and treated as distinct, new variables each time they are encountered. There are 
copies of the shaded part. 
Terms are abbreviated using their starting alphanumerical; $\mathtt{tr}$ stands for $\mathtt{tree}$,
$\mathtt{se}$ for $\mathtt{structure\_element}$,
$\mathtt{he}$ for $\mathtt{helical}$, $\mathtt{si}$ for $\mathtt{single}$, $\mathtt{nuc}$
for $\mathtt{nucleotide}$, and $\mathtt{nuc\_p}$ for $\mathtt{nucleotide\_pair}$.}
\end{center}
\end{figure}

\end{appendix}

\bibliographystyle{theapa}
\bibliography{kersting06a}

\begin{thebibliography}{}

\bibitem[\protect\BCAY{Abney}{Abney}{1997}]{Abney97d}
Abney, S. \BBOP1997\BBCP.
\newblock \BBOQ Stochastic {A}ttribute-{V}alue {G}rammars\BBCQ\
\newblock {\Bem {C}omputational {L}inguistics}, {\Bem 23\/}(4), 597--618.

\bibitem[\protect\BCAY{Abney, McAllester, \BBA\ Pereira}{Abney
  et~al.}{1999}]{AbneyMcAllesterPereira:99}
Abney, S., McAllester, D., \BBA\ Pereira, F. \BBOP1999\BBCP.
\newblock \BBOQ Relating probabilistic grammars and automata\BBCQ\
\newblock In {\Bem Proceedings of 37th Annual Meeting of the Association for
  Computational Linguistics (ACL-1999)}, \BPGS\ 542--549. Morgan Kaufmann.

\bibitem[\protect\BCAY{Anderson, Domingos, \BBA\ Weld}{Anderson
  et~al.}{2002}]{AndersonDomingosWeld:02}
Anderson, C., Domingos, P., \BBA\ Weld, D. \BBOP2002\BBCP.
\newblock \BBOQ {R}elational {M}arkov {M}odels and their {A}pplication to
  {A}daptive {W}eb {N}avigation\BBCQ\
\newblock In {\Bem Proceedings of the Eighth International Conference on
  Knowledge Discovery and Data Mining (KDD-2002)}, \BPGS\ 143--152\ Edmonton,
  Canada. ACM Press.

\bibitem[\protect\BCAY{Baker}{Baker}{1979}]{Baker:79}
Baker, J. \BBOP1979\BBCP.
\newblock \BBOQ Trainable grammars for speech recognition\BBCQ\
\newblock In {\Bem Speech communication paper presented at th 97th {M}eeting of
  the {A}coustical {S}ociety of {A}merica}, \BPGS\ 547--550\ Boston, MA.

\bibitem[\protect\BCAY{Bauer}{Bauer}{1991}]{Bauer:91}
Bauer, H. \BBOP1991\BBCP.
\newblock {\Bem Wahrscheinlichkeitstheorie\/} (4. \BEd).
\newblock Walter de {G}ruyter, Berlin, New York.

\bibitem[\protect\BCAY{Baum}{Baum}{1972}]{Baum:72}
Baum, L. \BBOP1972\BBCP.
\newblock \BBOQ An inequality and associated maximization technique in
  statistical estimation for probabilistic functions of markov processes\BBCQ\
\newblock {\Bem Inequalities}, {\Bem 3}, 1--8.

\bibitem[\protect\BCAY{Bohnebeck, Horv{\'a}th, \BBA\ Wrobel}{Bohnebeck
  et~al.}{1998}]{BohnebeckHorvathWrobel:98}
Bohnebeck, U., Horv{\'a}th, T., \BBA\ Wrobel, S. \BBOP1998\BBCP.
\newblock \BBOQ Term comparison in first-order similarity measures\BBCQ\
\newblock In {\Bem Proceedings of the Eigth International Conference on
  Inductive Logic Programming (ILP-98)}, \lowercase{\BVOL}\ 1446 of {\Bem
  LNCS}, \BPGS\ 65--79. Springer.

\bibitem[\protect\BCAY{Bresnan}{Bresnan}{2001}]{Bresnan:01}
Bresnan, J. \BBOP2001\BBCP.
\newblock {\Bem {L}exical-{F}unctional {S}yntax}.
\newblock Blackwell, Malden, MA.

\bibitem[\protect\BCAY{Carrasco, Oncina, \BBA\ {Calera-Rubio}}{Carrasco
  et~al.}{2001}]{CarrascoOncinaCaleraRubio:01}
Carrasco, R., Oncina, J., \BBA\ {Calera-Rubio}, J. \BBOP2001\BBCP.
\newblock \BBOQ Stochastic inference of regular tree languages\BBCQ\
\newblock {\Bem Machine Learning}, {\Bem 44\/}(1/2), 185--197.

\bibitem[\protect\BCAY{Chandonia, Hon, Walker, {Lo Conte}, P.Koehl, \BBA\
  Brenner}{Chandonia et~al.}{2004}]{astral:04}
Chandonia, J., Hon, G., Walker, N., {Lo Conte}, L., P.Koehl, \BBA\ Brenner, S.
  \BBOP2004\BBCP.
\newblock \BBOQ The {ASTRAL} compendium in $2004$\BBCQ\
\newblock {\Bem Nucleic Acids Research}, {\Bem 32}, D189--D192.

\bibitem[\protect\BCAY{Davison \BBA\ Hirsh}{Davison \BBA\
  Hirsh}{1998}]{DavisonHirsh:98}
Davison, B.\BBACOMMA\  \BBA\ Hirsh, H. \BBOP1998\BBCP.
\newblock \BBOQ {P}redicting {S}equences of {U}ser {A}ctions\BBCQ\
\newblock In {\Bem Predicting the Future: AI Approaches to Time-Series
  Analysis}, \BPGS\ 5--12. AAAI Press.

\bibitem[\protect\BCAY{{De Raedt} \BBA\ Kersting}{{De Raedt} \BBA\
  Kersting}{2003}]{DeRaedtKersting:03}
{De Raedt}, L.\BBACOMMA\  \BBA\ Kersting, K. \BBOP2003\BBCP.
\newblock \BBOQ {P}robabilistic {L}ogic {L}earning\BBCQ\
\newblock {\Bem ACM-SIGKDD Explorations: Special issue on Multi-Relational Data
  Mining}, {\Bem 5\/}(1), 31--48.

\bibitem[\protect\BCAY{{De Raedt} \BBA\ Kersting}{{De Raedt} \BBA\
  Kersting}{2004}]{DeRaedtKersting:04}
{De Raedt}, L.\BBACOMMA\  \BBA\ Kersting, K. \BBOP2004\BBCP.
\newblock \BBOQ {P}robabilistic {I}nductive {L}ogic {P}rogramming\BBCQ\
\newblock In Ben-David, S., Case, J., \BBA\ Maruoka, A.\BEDS, {\Bem Proceedings
  of the 15th International Conference on Algorithmic Learning Theory
  (ALT-2004)}, \lowercase{\BVOL}\ 3244 of {\Bem LNCS}, \BPGS\ 19--36\ Padova,
  Italy. Springer.

\bibitem[\protect\BCAY{Durbin, Eddy, Krogh, \BBA\ Mitchison}{Durbin
  et~al.}{1998}]{DurbinEddyKroghMitchison:98}
Durbin, R., Eddy, S., Krogh, A., \BBA\ Mitchison, G. \BBOP1998\BBCP.
\newblock {\Bem {B}iological sequence analysis: {P}robabilistic models of
  proteins and nucleic acids}.
\newblock Cambridge University Press.

\bibitem[\protect\BCAY{D\v{z}eroski \BBA\ Lavra\v{c}}{D\v{z}eroski \BBA\
  Lavra\v{c}}{2001}]{DzeroskiLavrac:01}
D\v{z}eroski, S.\BBACOMMA\  \BBA\ Lavra\v{c}, N.\BEDS. \BBOP2001\BBCP.
\newblock {\Bem Relational data mining}.
\newblock Springer-Verlag, Berlin.

\bibitem[\protect\BCAY{Eddy \BBA\ Durbin}{Eddy \BBA\
  Durbin}{1994}]{EddyDurbin:94}
Eddy, S.\BBACOMMA\  \BBA\ Durbin, R. \BBOP1994\BBCP.
\newblock \BBOQ {RNA} sequence analysis using covariance models\BBCQ\
\newblock {\Bem Nucleic Acids Res.}, {\Bem 22\/}(11), 2079--2088.

\bibitem[\protect\BCAY{Eisele}{Eisele}{1994}]{Eisele:94}
Eisele, A. \BBOP1994\BBCP.
\newblock \BBOQ {T}owards probabilistic extensions of contraint-based
  grammars\BBCQ\
\newblock In D{\"o}rne, J.\BED, {\Bem Computational Aspects of Constraint-Based
  Linguistics Decription-II}. DYNA-2 deliverable R1.2.B.

\bibitem[\protect\BCAY{Fine, Singer, \BBA\ Tishby}{Fine
  et~al.}{1998}]{FineSingerTishby:98}
Fine, S., Singer, Y., \BBA\ Tishby, N. \BBOP1998\BBCP.
\newblock \BBOQ The hierarchical hidden markov model: analysis and
  applications\BBCQ\
\newblock {\Bem Machine Learning}, {\Bem 32}, 41--62.

\bibitem[\protect\BCAY{Frasconi, Soda, \BBA\ Vullo}{Frasconi
  et~al.}{2002}]{FrasconiSodaVullo:02}
Frasconi, P., Soda, G., \BBA\ Vullo, A. \BBOP2002\BBCP.
\newblock \BBOQ Hidden markov models for text categorization in multi-page
  documents\BBCQ\
\newblock {\Bem Journal of Intelligent Information Systems}, {\Bem 18},
  195--217.

\bibitem[\protect\BCAY{Friedman, Getoor, Koller, \BBA\ Pfeffer}{Friedman
  et~al.}{1999}]{FriedmanGetoorKollerPfeffer:99}
Friedman, N., Getoor, L., Koller, D., \BBA\ Pfeffer, A. \BBOP1999\BBCP.
\newblock \BBOQ Learning probabilistic relational models\BBCQ\
\newblock In {\Bem Proceedings of Sixteenth International Joint Conference on
  Artificial Intelligence (IJCAI-1999)}, \BPGS\ 1300--1307. Morgan Kaufmann.

\bibitem[\protect\BCAY{Fristedt \BBA\ Gray}{Fristedt \BBA\
  Gray}{1997}]{FristedtGray:97}
Fristedt, B.\BBACOMMA\  \BBA\ Gray, L. \BBOP1997\BBCP.
\newblock {\Bem {A} {M}odern {A}pproach to {P}robability {T}heory}.
\newblock {P}robability and its applications. {B}irkh{\"a}user {B}oston.

\bibitem[\protect\BCAY{Ghahramani \BBA\ Jordan}{Ghahramani \BBA\
  Jordan}{1997}]{GhahramaniJordan:97}
Ghahramani, Z.\BBACOMMA\  \BBA\ Jordan, M. \BBOP1997\BBCP.
\newblock \BBOQ Factorial hidden {M}arkov models\BBCQ\
\newblock {\Bem Machine Learning}, {\Bem 29}, 245--273.

\bibitem[\protect\BCAY{Goodman}{Goodman}{1997}]{Goodman:97}
Goodman, J. \BBOP1997\BBCP.
\newblock \BBOQ Probabilistic feature grammars\BBCQ\
\newblock In {\Bem Proceedings of the Fifth International Workshop on Parsing
  Technologies (IWPT-97)}\ Boston, MA, USA.

\bibitem[\protect\BCAY{Greenberg}{Greenberg}{1988}]{Greenberg:88}
Greenberg, S. \BBOP1988\BBCP.
\newblock \BBOQ {U}sing {U}nix: collected traces of 168 users\BBCQ\
\newblock \BTR, Dept. of Computer Science, University of Calgary, Alberta.

\bibitem[\protect\BCAY{Hopcroft \BBA\ Ullman}{Hopcroft \BBA\
  Ullman}{1979}]{HopcroftUllman:79}
Hopcroft, J.\BBACOMMA\  \BBA\ Ullman, J. \BBOP1979\BBCP.
\newblock {\Bem {I}ntroduction to {A}utomata {T}heory, {L}anguages, and
  {C}omputation}.
\newblock Addison-Wesley Publishing Company.

\bibitem[\protect\BCAY{Horv{\'a}th, Wrobel, \BBA\ Bohnebeck}{Horv{\'a}th
  et~al.}{2001}]{HorvathWrobelBohnebeck:01}
Horv{\'a}th, T., Wrobel, S., \BBA\ Bohnebeck, U. \BBOP2001\BBCP.
\newblock \BBOQ {R}elational {I}nstance-{B}ased learning with {L}ists and
  {T}erms\BBCQ\
\newblock {\Bem Machine Learning}, {\Bem 43\/}(1/2), 53--80.

\bibitem[\protect\BCAY{Hubbard, Murzin, Brenner, \BBA\ Chotia}{Hubbard
  et~al.}{1997}]{HubbardMurzinBrennerChotia:97}
Hubbard, T., Murzin, A., Brenner, S., \BBA\ Chotia, C. \BBOP1997\BBCP.
\newblock \BBOQ {\textit{SCOP}}: a structural classification of proteins
  database\BBCQ\
\newblock {\Bem NAR}, {\Bem 27\/}(1), 236--239.

\bibitem[\protect\BCAY{Jacobs \BBA\ Blockeel}{Jacobs \BBA\
  Blockeel}{2001}]{JacobsBlockeel:01}
Jacobs, N.\BBACOMMA\  \BBA\ Blockeel, H. \BBOP2001\BBCP.
\newblock \BBOQ {T}he {L}earning {S}hell: {A}utomated {M}acro
  {C}onstruction\BBCQ\
\newblock In {\Bem User Modeling 2001}, \BPGS\ 34--43.

\bibitem[\protect\BCAY{Jaeger}{Jaeger}{1997}]{Jaeger:97}
Jaeger, M. \BBOP1997\BBCP.
\newblock \BBOQ Relational {B}ayesian networks\BBCQ\
\newblock In {\Bem Proceedings of the Thirteenth Conference on Uncertainty in
  Artificial Intelligence (UAI)}, \BPGS\ 266--273. Morgan Kaufmann.

\bibitem[\protect\BCAY{Katz}{Katz}{1987}]{Katz:87}
Katz, S. \BBOP1987\BBCP.
\newblock \BBOQ Estimation of probabilities from sparse data for hte language
  model component of a speech recognizer\BBCQ\
\newblock {\Bem {IEEE} Transactions on Acoustics, Speech, and Signal Processing
  (ASSP)}, {\Bem 35}, 400--401.

\bibitem[\protect\BCAY{Kersting \BBA\ {De Raedt}}{Kersting \BBA\ {De
  Raedt}}{2001a}]{KerstingDeRaedt:01b}
Kersting, K.\BBACOMMA\  \BBA\ {De Raedt}, L. \BBOP2001a\BBCP.
\newblock \BBOQ {A}daptive {B}ayesian {L}ogic {P}rograms\BBCQ\
\newblock In Rouveirol, C.\BBACOMMA\  \BBA\ Sebag, M.\BEDS, {\Bem Proceedings
  of the 11th International Conference on Inductive Logic Programming
  (ILP-01)}, \lowercase{\BVOL}\ 2157 of {\Bem LNAI}, \BPGS\ 118--131. Springer.

\bibitem[\protect\BCAY{Kersting \BBA\ {De Raedt}}{Kersting \BBA\ {De
  Raedt}}{2001b}]{KerstingDeRaedt:01}
Kersting, K.\BBACOMMA\  \BBA\ {De Raedt}, L. \BBOP2001b\BBCP.
\newblock \BBOQ Towards {C}ombining {I}nductive {L}ogic {P}rogramming with
  {B}ayesian {N}etworks\BBCQ\
\newblock In Rouveirol, C.\BBACOMMA\  \BBA\ Sebag, M.\BEDS, {\Bem Proceedings
  of the 11th International Conference on Inductive Logic Programming
  (ILP-01)}, \lowercase{\BVOL}\ 2157 of {\Bem LNAI}, \BPGS\ 118--131. Springer.

\bibitem[\protect\BCAY{Kersting \BBA\ Raiko}{Kersting \BBA\
  Raiko}{2005}]{Kersting05UAI}
Kersting, K.\BBACOMMA\  \BBA\ Raiko, T. \BBOP2005\BBCP.
\newblock \BBOQ '{S}ay {EM}' for {S}electing {P}robabilistic {M}odels for
  {L}ogical {S}equences\BBCQ\
\newblock In Bacchus, F.\BBACOMMA\  \BBA\ Jaakkola, T.\BEDS, {\Bem Proceedings
  of the 21st Conference on Uncertainty in Artificial Intelligence, UAI 2005},
  \BPGS\ 300--307\ Edinburgh, Scotland.

\bibitem[\protect\BCAY{Kersting, Raiko, Kramer, \BBA\ {De Raedt}}{Kersting
  et~al.}{2003}]{KerstingRaikoKramerDeRaedt:03}
Kersting, K., Raiko, T., Kramer, S., \BBA\ {De Raedt}, L. \BBOP2003\BBCP.
\newblock \BBOQ Towards discovering structural signatures of protein folds
  based on logical hidden markov models\BBCQ\
\newblock In Altman, R., Dunker, A., Hunter, L., Jung, T., \BBA\ Klein,
  T.\BEDS, {\Bem Proceedings of the Pacific Symposium on Biocomputing
  (PSB-03)}, \BPGS\ 192--203\ Kauai, Hawaii, USA. World Scientific.

\bibitem[\protect\BCAY{Koivisto, Kivioja, Mannila, Rastas, \BBA\
  Ukkonen}{Koivisto et~al.}{2004}]{KoivistoKiviojaMannilaRastasUkkonen:04}
Koivisto, M., Kivioja, T., Mannila, H., Rastas, P., \BBA\ Ukkonen, E.
  \BBOP2004\BBCP.
\newblock \BBOQ {H}idden {M}arkov {M}odelling {T}echniques for {H}aplotype
  {A}nalysis\BBCQ\
\newblock In Ben-David, S., Case, J., \BBA\ Maruoka, A.\BEDS, {\Bem Proceedings
  of 15th International Conference on Algorithmic Learning Theory (ALT-04)},
  \lowercase{\BVOL}\ 3244 of {\Bem LNCS}, \BPGS\ 37--52. Springer.

\bibitem[\protect\BCAY{Koivisto, Perola, Varilo, Hennah, Ekelund, Lukk,
  Peltonen, Ukkonen, \BBA\ Mannila}{Koivisto
  et~al.}{2002}]{KoivistoPerolaVariloHennahEkelundLukkPeltonenUkkonenMannila:0%
2}
Koivisto, M., Perola, M., Varilo, T., Hennah, W., Ekelund, J., Lukk, M.,
  Peltonen, L., Ukkonen, E., \BBA\ Mannila, H. \BBOP2002\BBCP.
\newblock \BBOQ An {MDL} method for finding haplotype blocks and for estimating
  the strength of haplotype block boundaries\BBCQ\
\newblock In Altman, R., Dunker, A., Hunter, L., Jung, T., \BBA\ Klein,
  T.\BEDS, {\Bem Proceedings of the Pacific Symposium on Biocomputing
  (PSB-02)}, \BPGS\ 502--513. World Scientific.

\bibitem[\protect\BCAY{Korvemaker \BBA\ Greiner}{Korvemaker \BBA\
  Greiner}{2000}]{KorvemakerGreiner:00}
Korvemaker, B.\BBACOMMA\  \BBA\ Greiner, R. \BBOP2000\BBCP.
\newblock \BBOQ {P}redicting {UNIX} command files: Adjusting to user
  patterns\BBCQ\
\newblock In {\Bem Adaptive {U}ser {I}nterfaces: {P}apers from the 2000 {AAAI}
  {S}pring {S}ymposium}, \BPGS\ 59--64.

\bibitem[\protect\BCAY{Kulp, Haussler, Reese, \BBA\ Eeckman}{Kulp
  et~al.}{1996}]{KulpHausslerReeseEeckman:96}
Kulp, D., Haussler, D., Reese, M., \BBA\ Eeckman, F. \BBOP1996\BBCP.
\newblock \BBOQ {A} {G}eneralized {H}idden {M}arkov {M}odel for the
  {R}ecognition of {H}uman {G}enes in {DNA}\BBCQ\
\newblock In States, D., Agarwal, P., Gaasterland, T., Hunter, L., \BBA\ Smith,
  R.\BEDS, {\Bem Proceedings of the Fourth International Conference on
  Intelligent Systems for Molecular Biology,(ISMB-96)}, \BPGS\ 134--142\ St.
  Louis, MO, USA. AAAI.

\bibitem[\protect\BCAY{Lane}{Lane}{1999}]{Lane:99}
Lane, T. \BBOP1999\BBCP.
\newblock \BBOQ {H}idden {M}arkov {M}odels for {H}uman/{C}omputer {I}nterface
  {M}odeling\BBCQ\
\newblock In Rudstr{\"o}m, {\r{A}}.\BED, {\Bem Proceedings of the IJCAI-99
  Workshop on Learning about Users}, \BPGS\ 35--44\ Stockholm, Sweden.

\bibitem[\protect\BCAY{Lari \BBA\ Young}{Lari \BBA\ Young}{1990}]{LariYoung:90}
Lari, K.\BBACOMMA\  \BBA\ Young, S. \BBOP1990\BBCP.
\newblock \BBOQ The estimation of stochastic context-free grammars using the
  inside-outside algorithm\BBCQ\
\newblock {\Bem Computer Speech and Language}, {\Bem 4}, 35--56.

\bibitem[\protect\BCAY{Levy \BBA\ Joshi}{Levy \BBA\ Joshi}{1978}]{LevyJoshi:78}
Levy, L.\BBACOMMA\  \BBA\ Joshi, A. \BBOP1978\BBCP.
\newblock \BBOQ Skeletal structural descriptions\BBCQ\
\newblock {\Bem Information and Control}, {\Bem 2\/}(2), 192--211.

\bibitem[\protect\BCAY{McLachlan \BBA\ Krishnan}{McLachlan \BBA\
  Krishnan}{1997}]{mclachlan97em}
McLachlan, G.\BBACOMMA\  \BBA\ Krishnan, T. \BBOP1997\BBCP.
\newblock {\Bem {T}he {EM} {A}lgorithm and {E}xtensions}.
\newblock Wiley, New York.

\bibitem[\protect\BCAY{Mitchell}{Mitchell}{1997}]{Mitchell:97}
Mitchell, T.~M. \BBOP1997\BBCP.
\newblock {\Bem Machine Learning}.
\newblock The McGraw-Hill Companies, Inc.

\bibitem[\protect\BCAY{Muggleton}{Muggleton}{1996}]{Muggleton:96}
Muggleton, S. \BBOP1996\BBCP.
\newblock \BBOQ Stochastic logic programs\BBCQ\
\newblock In {De Raedt}, L.\BED, {\Bem Advances in Inductive Logic
  Programming}, \BPGS\ 254--264. IOS Press.

\bibitem[\protect\BCAY{Muggleton \BBA\ {De Raedt}}{Muggleton \BBA\ {De
  Raedt}}{1994}]{MuggletonDeRaedt:94}
Muggleton, S.\BBACOMMA\  \BBA\ {De Raedt}, L. \BBOP1994\BBCP.
\newblock \BBOQ Inductive logic programming: Theory and methods\BBCQ\
\newblock {\Bem Journal of Logic Programming}, {\Bem 19\/}(20), 629--679.

\bibitem[\protect\BCAY{Ngo \BBA\ Haddawy}{Ngo \BBA\
  Haddawy}{1997}]{NgoHaddawy:97}
Ngo, L.\BBACOMMA\  \BBA\ Haddawy, P. \BBOP1997\BBCP.
\newblock \BBOQ Answering queries from context-sensitive probabilistic
  knowledge bases\BBCQ\
\newblock {\Bem Theoretical Computer Science}, {\Bem 171}, 147--177.

\bibitem[\protect\BCAY{Pollard \BBA\ Sag}{Pollard \BBA\
  Sag}{1994}]{PollardSag:94}
Pollard, C.\BBACOMMA\  \BBA\ Sag, I. \BBOP1994\BBCP.
\newblock {\Bem {H}ead-driven {P}hrase {S}tructure Grammar}.
\newblock The University of Chicago Press, Chicago.

\bibitem[\protect\BCAY{Rabiner \BBA\ Juang}{Rabiner \BBA\
  Juang}{1986}]{rabiner86introduction}
Rabiner, L.\BBACOMMA\  \BBA\ Juang, B. \BBOP1986\BBCP.
\newblock \BBOQ {A}n {I}ntroduction to {H}idden {M}arkov {M}odels\BBCQ\
\newblock {\Bem IEEE ASSP Magazine}, {\Bem 3\/}(1), 4--16.

\bibitem[\protect\BCAY{Riezler}{Riezler}{1998}]{Riezler:98}
Riezler, S. \BBOP1998\BBCP.
\newblock \BBOQ Statistical inference and probabilistic modelling for
  constraint-based nlp\BBCQ\
\newblock In Schröder, B., Lenders, W., \BBA\ und T.~Portele, W.~H.\BEDS,
  {\Bem Proceedings of the 4th Conference on Natural Language Processing
  (KONVENS-98)}.
\newblock Also as CoRR cs.CL/9905010.

\bibitem[\protect\BCAY{Sakakibara}{Sakakibara}{1992}]{Sakakibara:92}
Sakakibara, Y. \BBOP1992\BBCP.
\newblock \BBOQ Efficient learning of context-free grammars from positive
  structural examples\BBCQ\
\newblock {\Bem Information and Computation}, {\Bem 97\/}(1), 23--60.

\bibitem[\protect\BCAY{Sakakibara}{Sakakibara}{2003}]{Sakakibara:03}
Sakakibara, Y. \BBOP2003\BBCP.
\newblock \BBOQ Pair hidden markov models on tree structures\BBCQ\
\newblock {\Bem Bioinformatics}, {\Bem 19\/}(Suppl.1), i232--i240.

\bibitem[\protect\BCAY{Sakakibara, Brown, Hughey, Mian, Sjolander, \BBA\
  Underwood}{Sakakibara et~al.}{1994}]{SakakibaraBrown:94}
Sakakibara, Y., Brown, M., Hughey, R., Mian, I., Sjolander, K., \BBA\
  Underwood, R. \BBOP1994\BBCP.
\newblock \BBOQ Stochastic context-free grammars for {tRNA} modelling\BBCQ\
\newblock {\Bem Nucleic Acids Research}, {\Bem 22\/}(23), 5112--5120.

\bibitem[\protect\BCAY{Sanghai, Domingos, \BBA\ Weld}{Sanghai
  et~al.}{2003}]{SangheiDomingosWeld:03}
Sanghai, S., Domingos, P., \BBA\ Weld, D. \BBOP2003\BBCP.
\newblock \BBOQ Dynamic probabilistic relational models\BBCQ\
\newblock In Gottlob, G.\BBACOMMA\  \BBA\ Walsh, T.\BEDS, {\Bem Proceedings of
  the Eighteenth International Joint Conference on Artificial Intelligence
  (IJCAI-03)}, \BPGS\ 992--997\ Acapulco, Mexico. Morgan Kaufmann.

\bibitem[\protect\BCAY{{S}ato \BBA\ {K}ameya}{{S}ato \BBA\
  {K}ameya}{2001}]{Sato01parameterlearning}
{S}ato, T.\BBACOMMA\  \BBA\ {K}ameya, Y. \BBOP2001\BBCP.
\newblock \BBOQ Parameter learning of logic programs for symbolic-statistical
  modeling\BBCQ\
\newblock {\Bem Journal of Artificial Intelligence Research (JAIR)}, {\Bem 15},
  391--454.

\bibitem[\protect\BCAY{Johnson}{Sch{\"o}lkopf \BBA\ Warmuth}{2003}]{johnson:03}
Sch{\"o}lkopf, B.\BBACOMMA\  \BBA\ Warmuth, M.\BEDS. \BBOP2003\BBCP.
\newblock {\Bem Learning and Parsing Stochastic Unification-Based Grammars},
  \lowercase{\BVOL}\ 2777 of {\Bem LNCS}. Springer.

\bibitem[\protect\BCAY{Turcotte, Muggleton, \BBA\ Sternberg}{Turcotte
  et~al.}{2001}]{Turcotte_et_al_01a}
Turcotte, M., Muggleton, S., \BBA\ Sternberg, M. \BBOP2001\BBCP.
\newblock \BBOQ The effect of relational background knowledge on learning of
  protein three-dimensional fold signatures\BBCQ\
\newblock {\Bem Machine Learning}, {\Bem 43\/}(1/2), 81--95.

\bibitem[\protect\BCAY{Won, {Pr\"ugel-Bennett}, \BBA\ Krogh}{Won
  et~al.}{2004}]{WonPruegelBennettKrogh:04}
Won, K., {Pr\"ugel-Bennett}, A., \BBA\ Krogh, A. \BBOP2004\BBCP.
\newblock \BBOQ The {B}lock {H}idden {M}arkov {M}odel for {B}iological
  {S}equence {A}nalysis\BBCQ\
\newblock In Negoita, M., Howlett, R., \BBA\ Jain, L.\BEDS, {\Bem Proceedings
  of the Eighth International Conference on Knowledge-Based Intelligent
  Information and Engineering Systems (KES-04)}, \lowercase{\BVOL}\ 3213 of
  {\Bem LNCS}, \BPGS\ 64--70. Springer.

\end{thebibliography}


\begin{thebibliography}{}

\bibitem[\protect\BCAY{Aloul, Markov,\ \BBA\ Sakallah}{Aloul
  et~al.}{2003}]{Aloul03a}
Aloul, F.~A., Markov, I.~L., \BBA\ Sakallah, K.~A. \BBOP2003\BBCP.
\newblock \BBOQ Shatter: Efficient symmetry-breaking for boolean
  satisfiability\BBCQ\
\newblock In {\Bem International Joint Conference on Artificial Intelligence},
  \BPGS\ 271--282.

\bibitem[\protect\BCAY{Aloul, Markov,\ \BBA\ Sakallah}{Aloul
  et~al.}{2004}]{Aloul04b}
Aloul, F.~A., Markov, I.~L., \BBA\ Sakallah, K.~A. \BBOP2004\BBCP.
\newblock \BBOQ {MINCE}: A static global variable-ordering heuristic for sat
  search and bdd manipulation\BBCQ\
\newblock {\Bem Journal of Universal Computer Science (JUCS)}, {\Bem 10},
  1562--1596.

\bibitem[\protect\BCAY{Aloul, Ramani, Markov,\ \BBA\ Sakallah}{Aloul
  et~al.}{2002}]{Aloul02b}
Aloul, F.~A., Ramani, A., Markov, I.~L., \BBA\ Sakallah, K.~A. \BBOP2002\BBCP.
\newblock \BBOQ Generic {ILP} versus specialized 0-1 {ILP}: An update\BBCQ\
\newblock In {\Bem International Conference on Computer-Aided Design}, \BPGS\
  450--457.

\bibitem[\protect\BCAY{Aloul, Ramani, Markov,\ \BBA\ Sakallah}{Aloul
  et~al.}{2003}]{Aloul02a}
Aloul, F.~A., Ramani, A., Markov, I.~L., \BBA\ Sakallah, K.~A. \BBOP2003\BBCP.
\newblock \BBOQ Solving difficult instances of boolean satisfiability in the
  presence of symmetry\BBCQ\
\newblock {\Bem IEEE Transactions on CAD}, {\Bem 22}, 1117--1137.

\bibitem[\protect\BCAY{Aloul, Ramani, Markov,\ \BBA\ Sakallah}{Aloul
  et~al.}{2004}]{Aloul04}
Aloul, F.~A., Ramani, A., Markov, I.~L., \BBA\ Sakallah, K.~A. \BBOP2004\BBCP.
\newblock \BBOQ Symmetry-breaking for pseudo-boolean formulas\BBCQ\
\newblock In {\Bem Asia-Pacific Design Automation Conference}, \BPGS\ 884--887.

\bibitem[\protect\BCAY{Aragon, Johnson, McGeoch,\ \BBA\ Schevon}{Aragon
  et~al.}{1991}]{Johnson91}
Aragon, C.~R., Johnson, D.~S., McGeoch, L.~A., \BBA\ Schevon, C.
  \BBOP1991\BBCP.
\newblock \BBOQ Optimization by simulated annealing: An experimental
  evaluation; part ii, graph coloring and number partitioning\BBCQ\
\newblock {\Bem Operations Research}, {\Bem 39}, 378--406.

\bibitem[\protect\BCAY{Benhamou}{Benhamou}{2004}]{Benhamou04}
Benhamou, B. \BBOP2004\BBCP.
\newblock \BBOQ Symmetry in not-equals binary constraint networks\BBCQ\
\newblock In {\Bem Workshop on Symmetry in {CSP}s}, \BPGS\ 2--8.

\bibitem[\protect\BCAY{Brelaz}{Brelaz}{1979}]{Brelaz79}
Brelaz, D. \BBOP1979\BBCP.
\newblock \BBOQ New methods to color vertices of a graph\BBCQ\
\newblock {\Bem Communications of the ACM}, {\Bem 22}, 251--256.

\bibitem[\protect\BCAY{Brown}{Brown}{1972}]{Brown72}
Brown, R.~J. \BBOP1972\BBCP.
\newblock \BBOQ Chromatic scheduling and the chromatic number problem\BBCQ\
\newblock {\Bem Management Science}, {\Bem 19}, 451--463.

\bibitem[\protect\BCAY{Cadoli\ \BBA\ Mancini}{Cadoli\ \BBA\
  Mancini}{2003}]{Cadoli03}
Cadoli, M.\BBACOMMA\  \BBA\ Mancini, T. \BBOP2003\BBCP.
\newblock \BBOQ Detecting and breaking symmetries on specifications\BBCQ\
\newblock In {\Bem The Third Annual Workshop on Symmetry in Constraint
  Satisfaction Problems (SymCon)}, \BPGS\ 13--26.

\bibitem[\protect\BCAY{Cadoli, Palopoli, Schaerf,\ \BBA\ Vasileet}{Cadoli
  et~al.}{1999}]{Cadoli99}
Cadoli, M., Palopoli, L., Schaerf, A., \BBA\ Vasileet, D. \BBOP1999\BBCP.
\newblock \BBOQ {NP-SPEC}: An executable specification language for solving all
  problems in {NP}\BBCQ\
\newblock In {\Bem Practical Aspects of Declarative Languages}, \BPGS\ 16--30.

\bibitem[\protect\BCAY{Chai\ \BBA\ Kuehlmann}{Chai\ \BBA\
  Kuehlmann}{2003}]{Chai03}
Chai, D.\BBACOMMA\  \BBA\ Kuehlmann, A. \BBOP2003\BBCP.
\newblock \BBOQ A fast pseudo-boolean constraint solver\BBCQ\
\newblock In {\Bem Design Automation Conference}, \BPGS\ 830--835.

\bibitem[\protect\BCAY{Chaitin, Auslander, Chandra, Cocke, Hopkins,\ \BBA\
  Markstein}{Chaitin et~al.}{1981}]{Chaitin81}
Chaitin, G.~J., Auslander, M., Chandra, A., Cocke, J., Hopkins, M., \BBA\
  Markstein, P. \BBOP1981\BBCP.
\newblock \BBOQ Register allocation via coloring\BBCQ\
\newblock {\Bem Computer Languages}, {\Bem 6}, 47--57.

\bibitem[\protect\BCAY{Chams, Hertz,\ \BBA\ Werra}{Chams
  et~al.}{1987}]{Chams87}
Chams, M., Hertz, A., \BBA\ Werra, D.~D. \BBOP1987\BBCP.
\newblock \BBOQ Some experiments with simulated annealing for coloring
  graphs\BBCQ\
\newblock {\Bem European Journal of Operations Research}, {\Bem 32}, 260--266.

\bibitem[\protect\BCAY{Coudert}{Coudert}{1997}]{Coudert97}
Coudert, O. \BBOP1997\BBCP.
\newblock \BBOQ Coloring of real-life graphs is easy\BBCQ\
\newblock In {\Bem Design Automation Conference}, \BPGS\ 121--126.

\bibitem[\protect\BCAY{Crawford}{Crawford}{1992}]{Crawford92}
Crawford, J. \BBOP1992\BBCP.
\newblock \BBOQ A theoretical analysis of reasoning by symmetry in first-order
  logic\BBCQ\
\newblock In {\Bem AAAI Workshop on Tractable Reasoning at the Tenth National
  Conference on Artificial Intelligence}.

\bibitem[\protect\BCAY{Crawford, Ginsberg, Luks,\ \BBA\ Roy}{Crawford
  et~al.}{1996}]{Crawford96}
Crawford, J., Ginsberg, M., Luks, E., \BBA\ Roy, A. \BBOP1996\BBCP.
\newblock \BBOQ Symmetry-breaking predicates for search problems\BBCQ\
\newblock In {\Bem 5th International Conference on Principles of Knowledge
  Representation and Reasoning}, \BPGS\ 148--159.

\bibitem[\protect\BCAY{Culberson}{Culberson}{2004}]{Culberson}
Culberson, J. \BBOP2004\BBCP.
\newblock \BBOQ Graph coloring page\BBCQ\
\newblock {\tt \small http://web.cs.ualberta.ca/\~{ }joe/Coloring/index.html}.

\bibitem[\protect\BCAY{Darga, Liffiton, Sakallah,\ \BBA\ Markov}{Darga
  et~al.}{2004}]{Saucy}
Darga, P.~T., Liffiton, M.~H., Sakallah, K.~A., \BBA\ Markov, I.~L.
  \BBOP2004\BBCP.
\newblock \BBOQ Exploiting structure in symmetry generation for cnf\BBCQ\
\newblock In {\Bem 41st Internation Design Automation Conference}, \BPGS\
  530--534.

\bibitem[\protect\BCAY{Davis, Logemann,\ \BBA\ Loveland}{Davis
  et~al.}{1962}]{Davis62}
Davis, M., Logemann, G., \BBA\ Loveland, D. \BBOP1962\BBCP.
\newblock \BBOQ A machine program for theorem proving\BBCQ\
\newblock {\Bem Communications of the ACM}, {\Bem 5}, 394--397.

\bibitem[\protect\BCAY{Fahle, Schamberger,\ \BBA\ Sellmann}{Fahle
  et~al.}{2001}]{Fahle01}
Fahle, T., Schamberger, S., \BBA\ Sellmann, M. \BBOP2001\BBCP.
\newblock \BBOQ Symmetry breaking\BBCQ\
\newblock In {\Bem 7th International Conference on Principles and Practice of
  Constraint Programming}, \BPGS\ 93--107.

\bibitem[\protect\BCAY{Feige, Goldwasser, Lovasz, Safra,\ \BBA\ Szege}{Feige
  et~al.}{1991}]{Feige91}
Feige, U., Goldwasser, S., Lovasz, L., Safra, S., \BBA\ Szege, M.
  \BBOP1991\BBCP.
\newblock \BBOQ Approximating clique is almost {NP}-complete\BBCQ\
\newblock In {\Bem IEEE Symposium on Foundations of Computer Science}, \BPGS\
  2--12.

\bibitem[\protect\BCAY{Focacci\ \BBA\ Milano}{Focacci\ \BBA\
  Milano}{2001}]{Foc01}
Focacci, F.\BBACOMMA\  \BBA\ Milano, M. \BBOP2001\BBCP.
\newblock \BBOQ Global cut framework for removing symmetries\BBCQ\
\newblock In {\Bem Principles and Practice of Constraints Programming}, \BPGS\
  77--82.

\bibitem[\protect\BCAY{Galinier\ \BBA\ Hao}{Galinier\ \BBA\ Hao}{1999}]{Gal99}
Galinier, P.\BBACOMMA\  \BBA\ Hao, J. \BBOP1999\BBCP.
\newblock \BBOQ Hybrid evolutionary algorithms for graph coloring\BBCQ\
\newblock {\Bem Journal of Combinatorial Optimization}, {\Bem 3}, 379--397.

\bibitem[\protect\BCAY{Garey\ \BBA\ Johnson}{Garey\ \BBA\
  Johnson}{1979}]{GareyNP}
Garey, M.~R.\BBACOMMA\  \BBA\ Johnson, D.~S. \BBOP1979\BBCP.
\newblock {\Bem Computers and Intractability: A Guide to the Theory of
  {NP}-completeness}.
\newblock W. H. Freeman and Company.

\bibitem[\protect\BCAY{Gent}{Gent}{2001}]{Gent01}
Gent, I.~P. \BBOP2001\BBCP.
\newblock \BBOQ A symmetry-breaking constraint for indistinguishable
  values\BBCQ\
\newblock In {\Bem Workshop on Symmetry in Constraint Satisfaction Problems}.

\bibitem[\protect\BCAY{Goldberg\ \BBA\ Novikov}{Goldberg\ \BBA\
  Novikov}{2002}]{Berkmin}
Goldberg, E.\BBACOMMA\  \BBA\ Novikov, Y. \BBOP2002\BBCP.
\newblock \BBOQ Berkmin: A fast and robust {SAT}-solver\BBCQ\
\newblock In {\Bem Design Automation and Test in Europe}, \BPGS\ 142--149.

\bibitem[\protect\BCAY{Haldorsson}{Haldorsson}{1990}]{Hald90}
Haldorsson, M.~M. \BBOP1990\BBCP.
\newblock \BBOQ A still better performance guarantee for approximate graph
  coloring\BBCQ.

\bibitem[\protect\BCAY{Hentenryck, Agren, Flener,\ \BBA\ Pearson}{Hentenryck
  et~al.}{2003}]{Henten03}
Hentenryck, P.~V., Agren, M., Flener, P., \BBA\ Pearson, J. \BBOP2003\BBCP.
\newblock \BBOQ Tractable symmetry breaking for {CSP}s with interchangeable
  values\BBCQ\
\newblock In {\Bem The International Joint Conference on Artificial
  Intelligence (IJCAI)}.

\bibitem[\protect\BCAY{Hertz\ \BBA\ Werra}{Hertz\ \BBA\ Werra}{1987}]{Hertz87}
Hertz, A.\BBACOMMA\  \BBA\ Werra, D.~D. \BBOP1987\BBCP.
\newblock \BBOQ Using tabu search techniques for graph coloring\BBCQ\
\newblock {\Bem Computing}, {\Bem 39}, 345--351.

\bibitem[\protect\BCAY{Huang\ \BBA\ Darwiche}{Huang\ \BBA\
  Darwiche}{2003}]{Huang03}
Huang, J.\BBACOMMA\  \BBA\ Darwiche, A. \BBOP2003\BBCP.
\newblock \BBOQ A structure-based variable ordering heuristic for {SAT}\BBCQ\
\newblock In {\Bem The International Joint Conference on Artificial
  Intelligence}, \BPGS\ 1167--1172.

\bibitem[\protect\BCAY{ILOG}{ILOG}{2000}]{Cplex}
ILOG \BBOP2000\BBCP.
\newblock \BBOQ {ILOG} {CPLEX} {ILP} solver, version 7.0\BBCQ\
\newblock {\tt \small http://www.ilog.com/products/cplex/}.

\bibitem[\protect\BCAY{J.-P.~Hamiez}{J.-P.~Hamiez}{2001}]{Hamiez01}
J.-P.~Hamiez, J.-K.~H. \BBOP2001\BBCP.
\newblock \BBOQ Scatter search for graph coloring\BBCQ\
\newblock In {\Bem The 5th European Conference on Artificial Evolution}, \BPGS\
  168--179.

\bibitem[\protect\BCAY{Jagota}{Jagota}{1996}]{Jagota96}
Jagota, A. \BBOP1996\BBCP.
\newblock \BBOQ An adaptive, multiple restarts neural network algorithm for
  graph coloring\BBCQ\
\newblock {\Bem European Journal of Operational Research}, {\Bem 93}, 257--270.

\bibitem[\protect\BCAY{Kirovski\ \BBA\ Potkonjak}{Kirovski\ \BBA\
  Potkonjak}{1998}]{KirovskiP98}
Kirovski, D.\BBACOMMA\  \BBA\ Potkonjak, M. \BBOP1998\BBCP.
\newblock \BBOQ Efficient coloring of a large spectrum of graph\BBCQ\
\newblock In {\Bem Design Automation Conference}.

\bibitem[\protect\BCAY{Krishnamurthy}{Krishnamurthy}{1985}]{Krish85}
Krishnamurthy, B. \BBOP1985\BBCP.
\newblock \BBOQ Short proofs for tricky formulas\BBCQ\
\newblock {\Bem Acta Informatica}, {\Bem 22}, 327--337.

\bibitem[\protect\BCAY{Kubale\ \BBA\ Jackowski}{Kubale\ \BBA\
  Jackowski}{1985}]{Kubale85}
Kubale, M.\BBACOMMA\  \BBA\ Jackowski, B. \BBOP1985\BBCP.
\newblock \BBOQ A generalized implicit enumeration algorithm for graph
  coloring\BBCQ\
\newblock {\Bem Communications of the ACM}, {\Bem 28}, 412--418.

\bibitem[\protect\BCAY{Kubale\ \BBA\ Kusz}{Kubale\ \BBA\ Kusz}{1983}]{Kubale83}
Kubale, M.\BBACOMMA\  \BBA\ Kusz, E. \BBOP1983\BBCP.
\newblock \BBOQ Computational experience with implicit enumeration algorithms
  for graph coloring\BBCQ\
\newblock In {\Bem Proceedings of the WG'83 International Workshop on Graph
  Theoretic Concepts in Computer Science}, \BPGS\ 167--176.

\bibitem[\protect\BCAY{Leighton}{Leighton}{1979}]{Leighton79}
Leighton, F. \BBOP1979\BBCP.
\newblock \BBOQ A graph coloring algorithm for large scheduling problems\BBCQ\
\newblock {\Bem Journal of Research of the National Bureau of Standards}, {\Bem
  84}, 489--506.

\bibitem[\protect\BCAY{McKay}{McKay}{1990}]{Nauty}
McKay, B.~D. \BBOP1990\BBCP.
\newblock \BBOQ Nauty user's guide (version 1.5)\BBCQ\
\newblock {\tt \small http://cs.anu.edu.au/\~{}bdm/nauty/}.

\bibitem[\protect\BCAY{Mehrotra\ \BBA\ Trick}{Mehrotra\ \BBA\
  Trick}{1996}]{Meh96}
Mehrotra, A.\BBACOMMA\  \BBA\ Trick, M.~A. \BBOP1996\BBCP.
\newblock \BBOQ A column generation approach for graph coloring\BBCQ\
\newblock {\Bem INFORMS Journal on Computing}, {\Bem 8}, 344--354.

\bibitem[\protect\BCAY{Moskewicz, Madigan, Zhao, Zhang,\ \BBA\ Malik}{Moskewicz
  et~al.}{2001}]{Chaff}
Moskewicz, M., Madigan, C., Zhao, Y., Zhang, L., \BBA\ Malik, S.
  \BBOP2001\BBCP.
\newblock \BBOQ Chaff: Engineering an efficient sat solver\BBCQ\
\newblock In {\Bem Design Automation Conference}, \BPGS\ 530--535.

\bibitem[\protect\BCAY{Mycielski}{Mycielski}{1955}]{Myciel55}
Mycielski, J. \BBOP1955\BBCP.
\newblock \BBOQ Sur le coloriage des graphs\BBCQ\
\newblock {\Bem Colloqium Mathematicum}, {\Bem 3}, 161--162.

\bibitem[\protect\BCAY{Prestwich}{Prestwich}{2002}]{Prestwich02}
Prestwich, S. \BBOP2002\BBCP.
\newblock \BBOQ Supersymmetric modelling for local search\BBCQ\
\newblock In {\Bem SymCon: Workshop on Symmetries in CSPs}, \BPGS\ 21--28.

\bibitem[\protect\BCAY{Puget}{Puget}{2002}]{Puget02}
Puget, J. \BBOP2002\BBCP.
\newblock \BBOQ Symmetry breaking revisited\BBCQ\
\newblock In {\Bem Principles and Practice of Constraints Programming}, \BPGS\
  446--461.

\bibitem[\protect\BCAY{Ramani, Aloul, Markov,\ \BBA\ Sakallah}{Ramani
  et~al.}{2004}]{Ramani04}
Ramani, A., Aloul, F.~A., Markov, I.~L., \BBA\ Sakallah, K.~A. \BBOP2004\BBCP.
\newblock \BBOQ Breaking instance-independent symmetries in exact graph
  coloring\BBCQ\
\newblock In {\Bem Design Automation and Test in Europe}, \BPGS\ 324--329.

\bibitem[\protect\BCAY{Roney-Dougal, Gent, Kelsey,\ \BBA\ Linton}{Roney-Dougal
  et~al.}{2004}]{Roney04}
Roney-Dougal, C.~M., Gent, I.~P., Kelsey, T., \BBA\ Linton, S. \BBOP2004\BBCP.
\newblock \BBOQ Tractable symmetry breaking using restricted search trees\BBCQ\
\newblock In {\Bem European Conference on Artificial Intelligence}, \BPGS\
  211--215.

\bibitem[\protect\BCAY{Sheini}{Sheini}{2004}]{Pueblo}
Sheini, H. \BBOP2004\BBCP.
\newblock \BBOQ Pueblo 0-1 {ILP} solver\BBCQ\
\newblock {\tt \small http://www.eecs.umich.edu/\~{}hsheini/pueblo/}.

\bibitem[\protect\BCAY{Silva\ \BBA\ Sakallah}{Silva\ \BBA\
  Sakallah}{1999}]{Grasp}
Silva, J. P.~M.\BBACOMMA\  \BBA\ Sakallah, K.~A. \BBOP1999\BBCP.
\newblock \BBOQ {GRASP}: A new search algorithm for satisfiability\BBCQ\
\newblock {\Bem IEEE Transactions On Computers}, {\Bem 48}, 506--521.

\bibitem[\protect\BCAY{Trick}{Trick}{1996}]{Trick}
Trick, M. \BBOP1996\BBCP.
\newblock \BBOQ Network resources for coloring a graph\BBCQ\
\newblock {\tt \tiny http://mat.gsia.cmu.edu/COLOR/color.html}.

\bibitem[\protect\BCAY{Walsh}{Walsh}{2001}]{Walsh01}
Walsh, T. \BBOP2001\BBCP.
\newblock \BBOQ Search on high degree graphs\BBCQ\
\newblock In {\Bem 17th International Joint Conference on Artificial
  Intelligence}, \BPGS\ 266--271.

\bibitem[\protect\BCAY{Warners}{Warners}{1998}]{Warners98}
Warners, J.~P. \BBOP1998\BBCP.
\newblock \BBOQ A linear-time transformation of linear inequalities into
  conjunctive normal form\BBCQ\
\newblock {\Bem Information Processing Letters}, {\Bem 68}, 63--69.

\bibitem[\protect\BCAY{Welsh\ \BBA\ Powell}{Welsh\ \BBA\
  Powell}{1967}]{Welsh67}
Welsh, D. J.~A.\BBACOMMA\  \BBA\ Powell, M.~B. \BBOP1967\BBCP.
\newblock \BBOQ An upper bound on the chromatic number of a graph and its
  application to timetabling problems\BBCQ\
\newblock {\Bem Computer Journal}, {\Bem 10}, 85--86.

\bibitem[\protect\BCAY{Werra}{Werra}{1985}]{Dewerra85}
Werra, D.~D. \BBOP1985\BBCP.
\newblock \BBOQ An introduction to timetabling\BBCQ\
\newblock {\Bem European Journal of Operations Research}, {\Bem 19}, 151--162.

\end{thebibliography}

\end{document}